\documentclass[11pt,a4paper]{article}

\usepackage[T1]{fontenc}
\usepackage[utf8]{inputenc}
\usepackage[british]{babel}
\usepackage{amsmath,amssymb}
\usepackage{graphicx}
\usepackage{booktabs}
\usepackage{multirow}
\usepackage{array}
\usepackage{tabularx}
\usepackage{longtable}
\usepackage{float}
\usepackage{url}
\usepackage{xcolor}
\usepackage{csquotes}
\usepackage[natbibapa]{apacite}
\usepackage[margin=2.5cm]{geometry}
\usepackage{hyperref}
\hypersetup{colorlinks=true,linkcolor=black,citecolor=black,urlcolor=blue}

\newcommand{\dprime}{$d'$}

\title{CANDOR: A Multilingual Provenance-Documented Corpus \\ for Dysarthric Speech Research}

\author{
Bernard Muller$^{1}$ \quad LaVonne Roberts$^{1,2}$ \quad Leo Kaiya$^{2}$ \quad Antonio Ortiz Barra\~{n}\'{o}n$^{3}$ \\[0.6em]
$^{1}$The Scott-Morgan Foundation, Torquay, United Kingdom \\
$^{2}$SMF Labs, 28 rue du Colonel Pierre Avia, 75015 Paris, France \\
$^{3}$Tecnol\'{o}gico de Monterrey, Monterrey, Mexico \\[0.6em]
ORCID: Bernard Muller 0009-0004-7494-305X \quad LaVonne Roberts 0009-0008-2883-3764 \\
Leo Kaiya 0009-0001-8310-0319 \quad Antonio Ortiz Barra\~{n}\'{o}n 0009-0009-2690-9283 \\[0.6em]
Corresponding author: LaVonne Roberts (\href{mailto:lavonne@scottmorganfoundation.org}{lavonne@scottmorganfoundation.org})
}

\date{}

\begin{document}
\maketitle

\begin{abstract}
\noindent CANDOR assembles 61 corpora across 17 languages and five aetiological
classes plus a mixed/other group. Its register counts 2,057 speakers classified
as dysarthric and 19,538 healthy controls. The collection contains 4,469.7 hours
of audio, including 1,178.8 hours in the dysarthric subset. Counts use
speech-impairment labels where available and diagnosis labels otherwise. The
collection includes 732 dysarthric speakers across 14 non-English languages; 194
speakers use six languages unsupported by Qwen3-ASR, the model underlying its
non-English baselines. We document composition, chain of custody, licences and
transfer terms, while distinguishing corpus-level provenance from
participant-level consent records, which we hold for few constituents. Among 31
dysarthric-bearing corpora, open licences cover 29.0\% of corpora, 10.5\% of
speakers and 8.2\% of hours. Excluding the English Speech Accessibility Project
raises the open hour share to 31.2\%, showing its dependence on this
collection's composition. CANDOR cannot be redistributed as a single audio
resource; we specify public, restricted and conditional artefacts. Severity
labels available in 15 corpora do not provide a validated cross-corpus mapping.
Within the non-English ASR baselines, only Mandarin yields a real-speech
estimate meeting our evaluation criteria. It differs sharply from the synthesis
result and remains imprecise; an English challenge result uses a different model
and protocol. CANDOR provides a documented basis for research on
under-represented languages while making constraints on evaluation,
comparability and sharing explicit.
\end{abstract}


\section{Introduction}
\label{sec:intro}

Dysarthria --- a motor speech disorder arising from neurological injury or
disease --- affects speakers with amyotrophic lateral sclerosis (ALS),
Parkinson's disease (PD), cerebral palsy (CP), stroke, traumatic brain injury
and a long tail of rarer conditions. Automatic speech recognition degrades
sharply on dysarthric input \citep{derussis2019}, and the speakers most harmed
by that degradation are precisely those for whom speech technology would be most
consequential.

Progress on this problem is constrained less by modelling ideas than by data.
A recent systematic review of 48 speech- and language-disorder corpora
\citep{almarridi2026review} found that only three are multilingual, and that
every non-English language it identified is served by exactly one corpus. The
resources that define the field are correspondingly small, English, and
single-aetiology:
TORGO contributes 8 dysarthric speakers \citep{rudzicz2012}, UA-Speech 15
\citep{kim2008}, SSNCE Tamil 20 \citep{ssnce2021}. Larger resources exist ---
notably the Speech Accessibility Project \citep{zheng2025} --- but are English
and governed by restrictive transfer agreements. The consequence is a literature
in which cross-lingual claims rest on two or three languages, aetiology effects
are confounded with corpus effects, and severity is operationalised differently
in almost every paper.

This paper describes CANDOR, a corpus assembled to relieve that constraint, and
exists for three reasons.

\paragraph{A citable corpus description.} Every downstream study using CANDOR
--- ours and, we hope, others' --- currently re-summarises the corpus in its
methods section. This paper is the definitional reference those studies should
cite instead.

\paragraph{A breadth contribution.} CANDOR comprises 2,057 dysarthric speakers
across 15 of its 17 languages and five aetiology classes plus a mixed/other group
(Section~\ref{sec:comp}). Its long
tail is the unusual part: 194 dysarthric speakers in six languages --- Akan, Ga,
Luganda, Swahili, Tamil and Slovak --- that Qwen3-ASR, the multilingual ASR model
used for our baselines, does not support at all. Section~\ref{sec:related}
compares its published scope with that of Project Euphonia.

\paragraph{A governance contribution.} Many CANDOR constituents cannot be
redistributed under their source terms. We therefore document their
composition, licences, provenance and release conditions while source audio
remains under the originating agreements. Section~\ref{sec:openlic} quantifies
how licence restrictions are distributed across this collection and reports its
sensitivity to the Speech Accessibility Project. This account allows researchers
to identify the material they can obtain and the conditions governing its use.

CANDOR does \emph{not} provide several things. There is no validated
cross-corpus severity mapping; Section~\ref{sec:harm} characterises the problem
rather than claiming a solution. Healthy-control speakers outnumber dysarthric
speakers by roughly nine to one (2.8 to one in hours), and the control speech is
not task-matched. Six of seventeen
languages fall outside the ASR base model's coverage. Section~\ref{sec:limits} states these and other
limitations directly.

\subsection{Related work: existing disordered-speech resources}
\label{sec:related}

\paragraph{The landscape has been systematically surveyed.} Al-Marridi et
al.\ \citep{almarridi2026review} review corpora for speech and language
disorders (SLD) across thirteen disorder categories, screening 34,325 articles
returned by 143 queries over four bibliographic databases and analysing the 48
resulting corpora by target population, language, data modality and task domain.
Two of their findings bear directly on this paper. First, dysarthria is the
best-represented condition in the surveyed landscape, appearing in 14 of the 48
corpora, ahead of autism spectrum disorder (8) and developmental language
disorder (5). Second, and more consequentially for our purposes:

\begin{quote}
\enquote{Over half ($n = 26$) of the databases contain English data, with three
of these also containing data from other languages in addition to English.
Aside from these multilingual databases, the majority focus on one language
only. While there is a wide range of language representation across the
databases, each of these languages has only one database serving them.}
\citep{almarridi2026review}
\end{quote}

Across 48 systematically identified SLD corpora, then, three are multilingual at
all, and every non-English language is served by exactly one resource. This is
independent, peer-reviewed evidence for the English-dominant, single-language,
one-corpus-per-language structure of the field --- a characterisation we would
otherwise be asserting on our own authority, and which motivates the breadth
argument this paper makes.

\paragraph{Their diagnosis is the one CANDOR acts on.} The same review treats
availability as a first-class attribute of every corpus it catalogues, and
closes by recommending improved database accessibility and the establishment of
standardised protocols for dataset development --- consistent labelling,
metadata annotation and quality control. Those are precisely the two problems
Sections~\ref{sec:gov} and \ref{sec:harm} address. CANDOR acts on the diagnosis
by assembling and governing a corpus rather than by cataloguing what exists, and
the two activities are complementary: a survey establishes what the landscape
contains, which is a prerequisite for deciding what to build. We note that the
review's scope is considerably broader than ours --- it spans child language
impairment and autism alongside motor speech disorders --- so we cite it for its
characterisation of the field, not as support for any particular claim about
CANDOR's scale or coverage. We also note, as a scope observation rather than a
criticism, that Project Euphonia does not appear in the review; its inclusion
criteria centre on peer-reviewed database publications, and Euphonia's data has
never been released, so its absence follows predictably from the search
strategy. That is why this section covers both the systematic review and
Euphonia: neither alone describes the whole landscape.

\paragraph{Project Euphonia.} Project Euphonia reports approximately 3,000
speakers, more than 1.5 million utterances and five languages
\citep{martin2025euphonia}. Its reported non-English cohort comprises 132
speakers, whereas CANDOR's register includes 732 across 14 non-English
languages. The populations and counting conventions differ, so these figures
describe published scope rather than directly comparable clinical coverage.
CANDOR also reports language-by-aetiology composition (Table~\ref{tab:aet}).
Both collections carry restrictions on audio access. CANDOR documents
composition, provenance and access conditions and plans to release permitted
supporting artefacts (Sections~\ref{sec:gov} and~\ref{sec:access}). Some
constituent resources were supported or co-authored by Project Euphonia
researchers; that overlap is disclosed in the competing-interests statement.

Alongside these, the corpora that define the academic literature remain small
and single-aetiology --- TORGO at 8 dysarthric speakers \citep{rudzicz2012},
UA-Speech at 15 \citep{kim2008}, SSNCE Tamil at 20 \citep{ssnce2021} --- while
the Speech Accessibility Project \citep{zheng2025} supplies scale in English
under a restrictive transfer agreement. CANDOR incorporates all four; SAP alone
accounts for 1,236 of our 2,057 dysarthric speakers, so our relationship to it
is dependency rather than comparison, and as SAP releases the remainder of its
cohort a single English corpus will approach CANDOR's whole-corpus speaker
count.

\paragraph{CANDOR is not the first attempt to make disordered speech shareable.}
Federated sharing initiatives and research-infrastructure consortia have
addressed the same fragmentation for some years. The clearest example is DELAD
--- the Database Enterprise for Language And speech Disorders
\citep{delad} --- founded to let researchers share disordered-speech corpora
across institutional and national boundaries, with a multilingual scope reported
at 13 languages. CANDOR reports an assembled collection using a shared manifest
schema and documented processing and governance procedures. DELAD addresses
discovery and sharing across institutions. These approaches address different
stages of research access; harmonisation alone does not make severity measures,
tasks or evaluation sets comparable.

\paragraph{Known absences.} Several established disordered-speech corpora are
\emph{not} in CANDOR. The Nemours database of dysarthric speech, the Whitaker database, the QoLT
Korean dysarthric corpus and CUDYS are all absent, in each case for acquisition
or licensing reasons rather than by design. Their inclusion would extend
aetiology and language coverage, and we would welcome routes to them. This list
is not exhaustive: it names the corpora most likely to be asked about rather
than every resource CANDOR lacks. For a systematic account of the wider
landscape we refer the reader to \citet{almarridi2026review}.

\section{Corpus Composition}
\label{sec:comp}

\subsection{Scale}

Table~\ref{tab:headline} gives the headline figures. All are computed by summing
constituent-corpus records in the CANDOR licence audit register, snapshotted
2026-07-20, with the following corrections to that snapshot: hours for the
two CDLI Ga corpora are measured from the source audio, because the register rows
for those corpora repeat the duration of a single file; one single-participant
corpus has been withdrawn; an unofficial copy of a student project is excluded
(Section~\ref{sec:limits}); three subsets of a multi-site challenge bundle have
been removed; recordings that duplicate another recording in UGAkan and PC-GITA
are removed (Section~\ref{sec:limits}); and a single-speaker corpus of a
non-dysarthric speech disorder is excluded (see the counting conventions below).
Every count is recomputed from the corrected
records rather than read from the register's aggregate rows. Derived and synthetic pools are excluded throughout: their
``speakers'' are voice conversions or text-to-speech re-renderings of speakers
already counted in the real rows, and including them would double-count.
Derived-pool provenance is documented separately in
Section~\ref{sec:derived}.

\begin{table}[htbp]
\centering
\small
\caption{CANDOR headline scale, real (citable) corpora only. Derived and
synthetic pools are excluded --- see Section~\ref{sec:derived}.}
\label{tab:headline}
\begin{tabular}{lr}
\toprule
\textbf{Quantity} & \textbf{Value} \\
\midrule
Constituent corpora            & 61 \\
Languages                      & 17 \\
Dysarthric speakers (symptom-level where available) & 2,057 \\
Healthy-control speakers       & 19,538 \\
Total speakers                 & 21,595 \\
\midrule
Dysarthric utterances          & 709,902 \\
Healthy-control utterances     & 1,568,587 \\
Total utterances               & 2,278,489 \\
\midrule
Dysarthric hours               & 1,178.8 \\
Healthy-control hours          & 3,290.8 \\
Total hours                    & 4,469.7 \\
\bottomrule
\end{tabular}
\end{table}

Of the 61 corpora, 31 contribute dysarthric speakers and 30 are healthy-control
resources included for language-matched balance.

\paragraph{Counting conventions.} CANDOR uses symptom-level counts where the
originating corpus provides them and diagnosis-level counts otherwise. A
diagnosis associated with dysarthria does not establish measurable speech
impairment in every participant. The headline total therefore combines counting
conventions. The convention used is recorded per corpus in the register; the
manifest's Boolean \texttt{is\_dysarthric} field does not preserve that
distinction. Cross-corpus counts should be interpreted with this limitation in
view.

VOC-ALS is where the distinction becomes concrete, and applying our own
convention to it is what fixes the headline figure. That corpus records
\textbf{102} speakers with an ALS \emph{diagnosis}, of whom \textbf{81}
\emph{present} dysarthria on its own Cantagallo self-assessment (score $> 6$);
the remaining 21 have no measurable speech impairment and are grouped with the
controls by the corpus authors. CANDOR counts these 21 in neither speaker column:
they are not among the dysarthric speakers and not among the healthy controls.
Their 168 recordings (0.8 hours) follow the corpus authors' grouping and are
included in the healthy-control utterance and hour totals of
Table~\ref{tab:headline}. The diagnosis-based and symptom-based
classifications describe the same 153 participants and 1,224 utterances. Using
VOC-ALS's symptom-based classification gives the headline total of 2,057.
Including its 21 additional participants with an ALS diagnosis would raise that
total to 2,078; this alternative does not establish a uniform diagnosis-based
count across the other corpora.

COPAS is counted from the pathology field of its source metadata: 75 speakers
labelled dysarthric (4,667 utterances, 2.1 hours) and 46 labelled normal (3,032
utterances, 1.1 hours). Its other 106 speakers have non-dysarthric pathologies
(cleft palate 37, hearing impairment 29, laryngectomy 30, voice disorder 7,
articulation disorder 2, unlabelled 1) and are excluded from both speaker
columns and from every utterance and hour total. The phonological \dprime{}
studies of Section~\ref{sec:base} include these other-pathology speakers (all 227
COPAS speakers in \citet{muller2026paper2}), so their COPAS counts differ from
those reported here.

The same rule excludes the single-speaker Japanese corpus SS-JDSC
\citep{ogasawara2026ssjdsc}: its one speaker has velopharyngeal insufficiency,
which is not a dysarthria, so the corpus (11,502 utterances, 15.1 hours) is
counted in neither speaker column and in no utterance or hour total. Japanese is
therefore represented by healthy-control corpora only.

\paragraph{Register and manifest counts.} Table~\ref{tab:headline} reports the
corrected 2026-07-20 register inventory; Table~\ref{tab:aet} reports speakers
represented in the processed manifests. Their totals are 2,057 and 2,055
respectively. The difference comprises three SAP speakers and one Domotica
speaker excluded because they have no usable audio, offset by one German and one
French speaker classified as dysarthric in the manifests but as controls in the
register. The first two differences arise from processing exclusions; the latter
two are register misclassifications. Headline scale uses the register basis,
while aetiology-specific statements use the manifest basis. The two register
corrections have not been incorporated into the headline total in this version.

VOC-ALS is visible here only because its severity scores are published at
speaker level; most corpora do not publish the per-speaker data that would let us
make the same check, so we neither claim they are free of the same gap nor assert
that they contain it (Section~\ref{sec:limits}).

The dysarthric constituents are TORGO \citep{rudzicz2012}, UA-Speech
\citep{kim2008}, the Speech Accessibility Project \citep{zheng2025}, XRMB
\citep{weismer2003xrmb}, HeyJay! \citep{morovelazquez2026heyjay}, Domotica and
Treasure Hunters \citep{ganzeboom2018}, CHASING \citep{ganzeboom2022chasing},
COPAS \citep{vannuffelen2009copas}, EasyCall \citep{turrisi2021easycall},
VOC-ALS \citep{dubbioso2024vocals}, IPVS \citep{dimauro2017ipvs}, SSNCE Tamil
\citep{ssnce2021}, MDSC
\citep{gao2024mdsc}, CDSD \citep{wan2024cdsd}, the Saarbr\"ucken Voice Database
\citep{puetzer2007svd}, AVFAD \citep{jesus2017avfad}, PC-GITA
\citep{orozco2014pcgita}, NeuroVoz \citep{mendes2024neurovoz}, the Hungarian
Dysarthria Database \citep{jenei2021hungarian}, EWA-DB \citep{rusko2024ewadb},
UGAkan \citep{wiafe2025ugakan} and the CDLI non-standard speech collection
\citep{cdli2024}. Healthy-control balance draws on
LibriSpeech \citep{panayotov2015}, VoxPopuli \citep{wang2021voxpopuli}, Common
Voice \citep{ardila2020commonvoice}, MLS \citep{pratap2020mls}, FLEURS
\citep{conneau2023fleurs}, AISHELL-1 \citep{bu2017aishell}, JVS
\citep{takamichi2020jvs}, CGN \citep{oostdijk2000cgn}, WAXAL
\citep{diack2026waxal} and the Google Indic corpora \citep{he2020slr65}. Full
per-corpus licence, custodian and citation detail is in the accompanying
register.

\subsection{Language distribution}

Table~\ref{tab:lang} gives the per-language composition. The distribution is
severely skewed: English holds 1,325 of 2,057 dysarthric speakers (64.4\%), and
a single corpus --- the Speech Accessibility Project --- holds 1,236 of those.
Excluding SAP, CANDOR comprises 821 dysarthric speakers, and English drops to
89 of them. SAP is a
transformative resource, but it is one corpus, in one language, under one
agreement, and conclusions that depend on it are not cross-lingual conclusions.

The tail is where the resource is unusual. Four African languages contribute 79
dysarthric speakers between them, Slovak contributes 95 in a single corpus, and
Hungarian, German and Mandarin each contribute cohorts in the 26--75
range. Six languages --- French, Portuguese, Luganda, Ga, Swahili and
Tamil --- contribute fewer than 25 dysarthric speakers each and should be treated as
pilot-scale. Czech and Japanese are represented by healthy-control corpora only.

\begin{table}[htbp]
\centering
\small
\caption{Per-language composition, real corpora only. Ordered by dysarthric
speaker count. Corpora column counts constituent corpora contributing to that
language, including healthy-control-only resources.}
\label{tab:lang}
\setlength{\tabcolsep}{4pt}
\begin{tabular}{lrrrrr}
\toprule
\textbf{Language} & \textbf{Corp.} & \textbf{Dys spk} & \textbf{HC spk} & \textbf{Dys h} & \textbf{HC h} \\
\midrule
English   & 9 & 1,325 & 3,023 & 912.1 & 1,485.2 \\
Italian   & 6 & 140 & 321 & 13.0 & 104.4 \\
Dutch     & 7 & 133 & 205 & 23.9 & 44.4 \\
Spanish   & 4 & 106 & 197 & 3.6 & 21.9 \\
Slovak    & 1 & 95 & 863 & 14.7 & 127.8 \\
Mandarin  & 5 & 75 & 673 & 114.2 & 265.1 \\
Hungarian & 6 & 48 & 1,720 & 12.5 & 204.6 \\
Akan      & 2 & 38 & 134 & 23.2 & 69.5 \\
German    & 3 & 26 & 216 & 1.2 & 19.2 \\
Swahili   & 1 & 24 & 1 & 17.8 & 0.9 \\
Tamil     & 3 & 20 & 63 & 12.0 & 21.8 \\
Ga        & 2 & 10 & 84 & 23.7 & 101.2 \\
Luganda   & 1 & 7 & 0 & 5.5 & 0.0 \\
Portuguese& 4 & 7 & 3,689 & 0.4 & 360.4 \\
French    & 2 & 3 & 163 & 1.0 & 13.6 \\
Czech     & 3 & 0 & 1,158 & 0.0 & 80.8 \\
Japanese  & 2 & 0 & 7,028 & 0.0 & 370.1 \\
\midrule
\textbf{Total} & \textbf{61} & \textbf{2,057} & \textbf{19,538} & \textbf{1,178.8} & \textbf{3,290.8} \\
\bottomrule
\end{tabular}
\end{table}

\subsection{Aetiology distribution}
\label{sec:aet}

\begin{figure}[htbp]
\centering
\includegraphics[width=\linewidth]{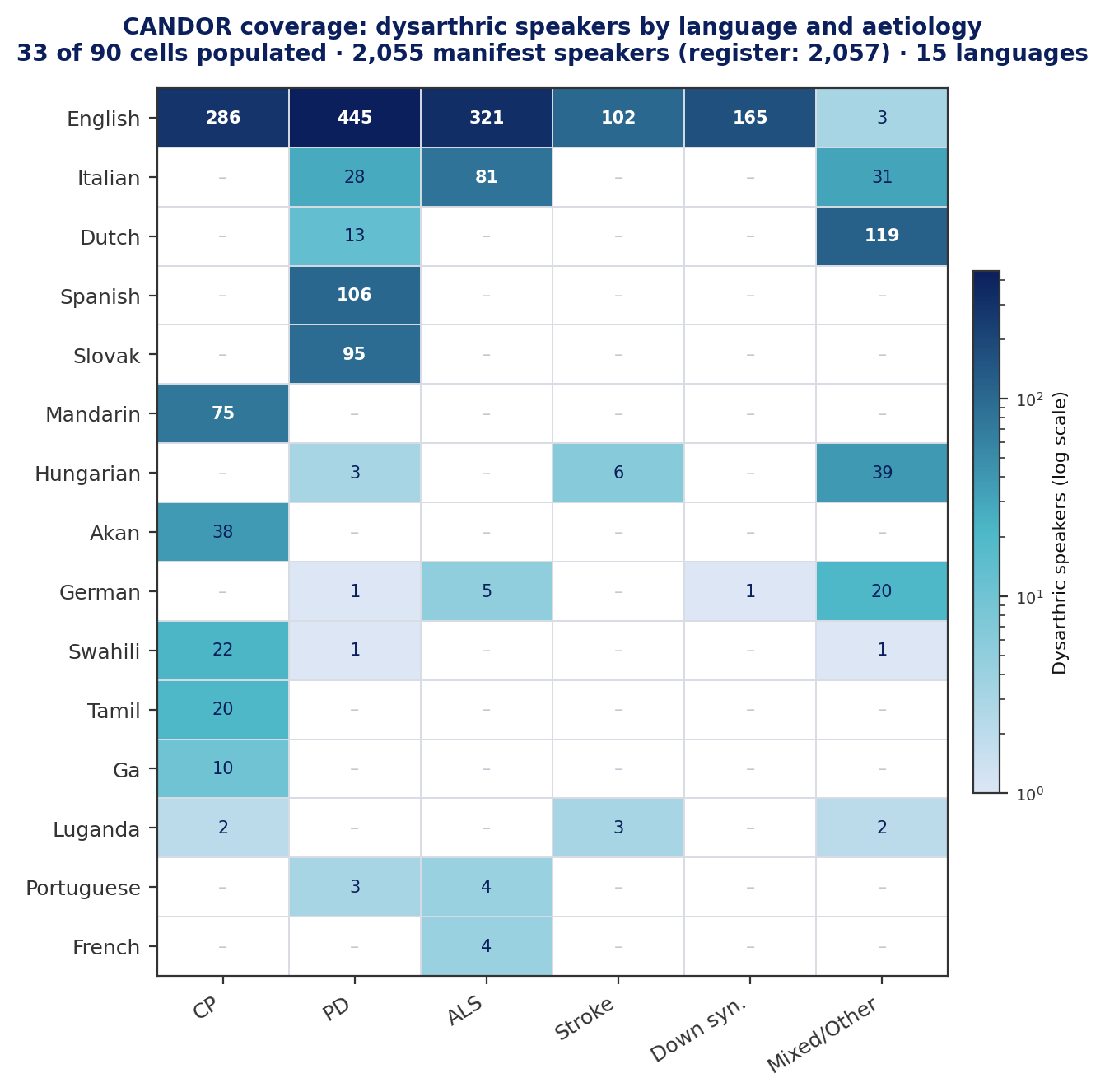}
\caption{CANDOR coverage: dysarthric speakers by language and aetiology, log
shading. Thirty-three of 90 cells are populated. Counts are manifest-derived and total
2,055 against the 2,057 of Table~\ref{tab:headline}: four register speakers have no
usable audio in the manifests and two speakers the register lists as controls are
dysarthric in the manifests (Section~\ref{sec:comp}). Cerebral palsy is the most
cross-lingually distributed aetiology (7 languages, including all four African
ones); Parkinson's dominates by count but is almost entirely European (694 of
695 speakers, the exception being one Swahili speaker); Down Syndrome and Stroke
are effectively single-corpus. The blank space is the
substantive content of the figure.}
\label{fig:matrix}
\end{figure}

Aetiology is recorded as free text by originating custodians, in forms ranging
from \enquote{CP (spastic)} to \enquote{Mixed dysarthria (PD, CVA, TBI, MS) + HC
+ other pathology}. We do not parse those strings. CANDOR's processing pipeline
writes a normalised per-record \texttt{aetiology} field at ingestion, and the
distribution reported here is derived from the manifests themselves --- counting
distinct dysarthric speaker identifiers across every real corpus --- rather than
from the register's free-text column. No speaker carries conflicting aetiology
labels across their records. Table~\ref{tab:aet} and Figure~\ref{fig:matrix}
give the resulting language $\times$ aetiology matrix.

\begin{table}[htbp]
\centering
\small
\caption{Language $\times$ aetiology matrix: distinct dysarthric speakers,
derived from normalised manifest records. Dashes mark empty cells. Mixed/Other
aggregates \texttt{mixed}, \texttt{unknown}, \texttt{voice\_disorder},
\texttt{neurodegenerative}, \texttt{multiple\_sclerosis} and
\texttt{hydrocephalus}. French counts four dysarthric speakers here and three in
Table~\ref{tab:lang}, and German 27 and 26: in each language one speaker is
labelled dysarthric in the manifests and as a healthy control in the register.
Both counts are reported as they stand.}
\label{tab:aet}
\setlength{\tabcolsep}{3.5pt}
\begin{tabular}{lrrrrrrr}
\toprule
\textbf{Language} & \textbf{CP} & \textbf{PD} & \textbf{ALS} & \textbf{Str.} & \textbf{DS} & \textbf{Mix.} & \textbf{Total} \\
\midrule
English    & 286 & 445 & 321 & 102 & 165 & 3   & \textbf{1,322} \\
Spanish    & --  & 106 & --  & --  & --  & --  & 106 \\
Italian    & --  & 28  & 81  & --  & --  & 31  & 140 \\
Dutch      & --  & 13  & --  & --  & --  & 119 & 132 \\
German     & --  & 1   & 5   & --  & 1   & 20  & 27 \\
Slovak     & --  & 95  & --  & --  & --  & --  & 95 \\
Mandarin   & 75  & --  & --  & --  & --  & --  & 75 \\
Hungarian  & --  & 3   & --  & 6   & --  & 39  & 48 \\
Akan       & 38  & --  & --  & --  & --  & --  & 38 \\
Swahili    & 22  & 1   & --  & --  & --  & 1   & 24 \\
Tamil      & 20  & --  & --  & --  & --  & --  & 20 \\
Ga         & 10  & --  & --  & --  & --  & --  & 10 \\
Luganda    & 2   & --  & --  & 3   & --  & 2   & 7 \\
Portuguese & --  & 3   & 4   & --  & --  & --  & 7 \\
French     & --  & --  & 4   & --  & --  & --  & 4 \\
\midrule
\textbf{Total} & \textbf{453} & \textbf{695} & \textbf{415} & \textbf{111} & \textbf{166} & \textbf{215} & \textbf{2,055} \\
\bottomrule
\end{tabular}
\end{table}

Four features deserve emphasis.

\paragraph{The largest corpus is aetiology-resolved.} SAP supplies per-speaker
diagnoses --- PD 420, ALS 311, CP 240, Down Syndrome 165, Stroke 97 --- and this
is what makes Figure~\ref{fig:matrix} possible at all. Had SAP been recorded only
as \enquote{mixed}, 60.0\% of the corpus would have collapsed into a single
undifferentiated cell and no aetiology-conditioned analysis would be available at
scale. This is not obvious from the corpus's public description and is worth
stating explicitly.

\paragraph{The matrix is sparse.} Thirty-three of 90 cells are populated. We report
the sparsity rather than smoothing over it: it is a more useful statement about
the state of the field than a filled table would be.

\paragraph{Cerebral palsy is the cross-lingual backbone.} CP spans seven
languages, including all four African ones, and is the strongest available basis
for cross-lingual claims in CANDOR. Parkinson's disease dominates by speaker
count (695) but is almost entirely confined to European languages: English,
Spanish, Slovak, German, Italian, Dutch, Hungarian and Portuguese account
for 694 of the 695, and a single Swahili speaker is the only non-European PD case
in the corpus. There is no South Asian PD cohort at all.

\paragraph{Two aetiologies are single-corpus.} Down Syndrome (166 speakers, 165
from SAP English) and Stroke (111 speakers, 97 from SAP and 102 English in total)
are effectively one corpus in one language each. We return to this in
Section~\ref{sec:limits}.

\subsection{Elicitation tasks}

Constituent corpora differ substantially in what speakers were asked to
produce. Read sentences are the modal task (26 corpora), followed by spontaneous
speech (14). A substantial minority elicit only isolated words (UA-Speech),
only short commands (Domotica, EasyCall), or only sustained vowels and
diadochokinetic repetitions (VOC-ALS). This heterogeneity is consequential and
under-reported: a model trained on command corpora underperforms on
conversational speech, and vowel-and-DDK corpora such as VOC-ALS do not contain
continuous speech at all and cannot support ASR evaluation. We record task type
per corpus in the register, and the planned public register carries it, so that downstream users can filter
appropriately rather than discovering the mismatch empirically.

\subsection{Severity label coverage}

Severity labels are the scarcest resource in the corpus. Of 61 corpora, 14 carry
severity labels covering 423 dysarthric speakers; one (SAP) is partially
labelled, with 188 of 1,236 speakers rated; and 46 carry none. Coverage is
therefore 423 fully-labelled speakers out of 2,057, or 611 including the SAP
partial subset. The instruments used are not commensurable, and we treat this as
an open methodological problem in Section~\ref{sec:harm} rather than as a
solved mapping.

\section{Governance Model}
\label{sec:gov}

\subsection{Licensing and clinical coverage in CANDOR}
\label{sec:openlic}

CANDOR's constituent corpora were acquired under 24 distinct licence types,
ranging from CC0 public-domain dedication through to bespoke bilateral data
sharing agreements. Grouping these by
redistribution posture gives Table~\ref{tab:redist}.

\paragraph{The denominator has to be the dysarthric-bearing corpora.} Computed
over all 61 constituent corpora, 66\% carry open licences under the loose
classification that counts non-commercial and no-derivatives terms as open, and
61\% under the classification used here. Either way the
denominator is wrong and the figure analytically misleading. Thirty of the 61
corpora are healthy-control resources contributing zero dysarthric speakers,
and 28 of those 30 are openly licensed --- because control
speech is LibriSpeech, VoxPopuli, Common Voice, MLS, FLEURS and AISHELL, which
are open almost by definition. Including them inflates the ``open'' share with
corpora that could not have carried clinical content in the first place. The
analysis below is therefore restricted to the 31 corpora that actually contribute
dysarthric speakers.

\paragraph{Non-commercial and no-derivatives licences are not open.} Neither
CC-BY-NC nor CC-BY-NC-ND satisfies any standard definition of an open licence --- one forbids
commercial use, and the other permits adapted material for non-commercial use
only and forbids sharing it (Section~\ref{sec:derived}). They are reported as their own restriction classes below.

\begin{table}[htbp]
\centering
\small
\caption{Redistribution posture across the 31 constituent corpora that
contribute dysarthric speakers. Percentages are of the column total. Without
SAP (30 corpora, 821 speakers, 310.6 hours) the open share is 30.0\% of corpora,
26.2\% of dysarthric speakers and 31.2\% of dysarthric hours.}
\label{tab:redist}
\setlength{\tabcolsep}{4pt}
\begin{tabular}{p{0.40\linewidth}rrr}
\toprule
\textbf{Posture} & \textbf{Corpora} & \textbf{Dys spk} & \textbf{Dys h} \\
\midrule
Open (CC0, CC-BY, CC-BY-SA)        & 9 (29.0\%)  & 215 (10.5\%)  & 97.0 (8.2\%) \\
Non-commercial (CC-BY-NC)          & 2 (6.5\%)   & 62 (3.0\%)   & 23.6 (2.0\%) \\
No-derivatives (CC-BY-NC-ND)       & 1 (3.2\%)   & 7 (0.3\%)    & 0.4 (0.0\%) \\
Signed agreement or custodian-set research-only conditions, no redistribution& 16 (51.6\%) & 1,678 (81.6\%)& 956.4 (81.1\%) \\
Redistribution explicitly prohibited& 3 (9.7\%) & 95 (4.6\%) & 101.6 (8.6\%) \\
\midrule
\textbf{Total} & \textbf{31} & \textbf{2,057} & \textbf{1,178.8} \\
\bottomrule
\end{tabular}
\end{table}

\begin{figure}[htbp]
\centering
\includegraphics[width=\linewidth]{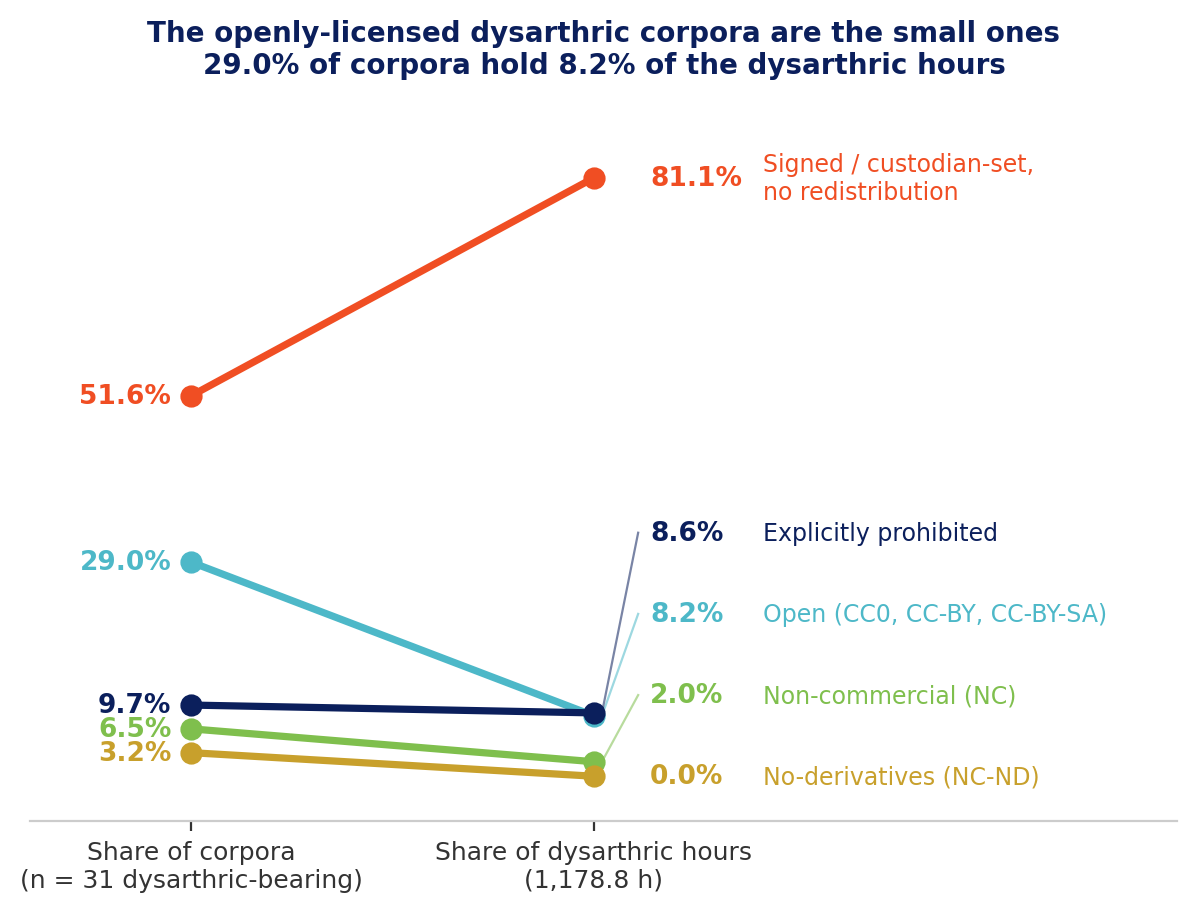}
\caption{Licence distribution across CANDOR's dysarthric-bearing corpora. Across the 31
constituent corpora that contribute dysarthric speakers, genuinely open licences
(CC0, CC-BY, CC-BY-SA) cover 29.0\% of corpora but only 10.5\% of dysarthric speakers
and 8.2\% of dysarthric hours --- a 20.8-point gap between corpus share and hour
share. Non-commercial and no-derivatives licences are shown separately from open
licences because each restricts what may be done with the derived artefacts
described in Section~\ref{sec:derived}.}
\label{fig:inversion}
\end{figure}

Nine of the 31 dysarthric-bearing corpora carry genuinely open licences:
CDLI Ga, CDLI Kenyan English, CDLI Kenyan Swahili, CDLI Luganda, CDLI Ugandan
English, IPVS, NeuroVoz, the Saarbr\"ucken Voice Database and UGAkan.
They are 29.0\% of the corpora, 10.5\% of the dysarthric speakers and 8.2\% of the
dysarthric hours (Figure~\ref{fig:inversion}). Open licensing and clinical
content are not structurally anti-correlated here --- nine open clinical corpora
exist and several are recent --- but the open corpora are small: the median
dysarthric-speaker count in the open group is in the tens, and the four largest clinical cohorts in CANDOR
(SAP, VOC-ALS, EWA-DB, COPAS) are all held under restrictive
terms.

Within this CANDOR snapshot, the open-licence subset contains 10.5\% of
dysarthric speakers and 8.2\% of dysarthric audio hours. Excluding SAP raises
these shares to 26.2\% and 31.2\%. The result therefore depends strongly on the
composition of this collection. It describes the coverage retained by selecting
CANDOR's openly licensed subset, not the proportion of all clinical speech
available to researchers. Publishing composition and access conditions makes
that selection visible even where the underlying audio cannot be redistributed.

\subsection{Licence taxonomy}

The 24 licence types fall into four functional families:

\begin{enumerate}
\item \textbf{Public-domain and permissive} (CC0, CC0-1.0, Apache-2.0; 13
corpora). Unrestricted redistribution. Entirely healthy-control: these corpora
contribute zero dysarthric speakers.
\item \textbf{Attribution-based Creative Commons} (CC-BY-4.0, CC-BY-SA-4.0,
CC-BY-NC-4.0, CC-BY-NC-ND-4.0; 28 corpora). Redistribution
permitted under attribution. Only the CC-BY and CC-BY-SA members are included in
the open category used in Section~\ref{sec:openlic}. The NC variants restrict
commercial use. The NC-ND variant distinguishes producing adapted material for
non-commercial use from sharing that material, as described in
Section~\ref{sec:derived}; we therefore report it separately. This
family carries all of CANDOR's openly licensed dysarthric content --- the five
dysarthric-bearing CDLI African corpora, UGAkan, IPVS, NeuroVoz and the
Saarbr\"ucken Voice Database --- plus three corpora (EasyCall and MDSC under NC; AVFAD under
NC-ND) that are attribution-licensed but not open.
\item \textbf{Institutional academic agreements and equivalent custodian-set research-only conditions} (Academic-DUA, LDC-DUA,
ICPSR Restricted-Use DUA, Custom-TDUA-UIUC, EULA, Restricted-professor-DUA,
Academic-non-commercial, Custom-licence-INT-Leiden, Custom-academic-licence, and public-repository custodian-set research-only conditions;
13 corpora). Signed or custodian-set, auditable, research-only, no redistribution. This family
carries the bulk of CANDOR's clinical mass, including SAP (1,236 speakers),
EWA-DB (95), Hungarian Dysarthria Database (47), CDSD (44), TORGO, UA-Speech,
SSNCE Tamil, HeyJay!\ and VOC-ALS (81 dysarthric speakers, accessed via Synapse under the custodian's terms: non-commercial research only, no redistribution).
\item \textbf{Informal and residual} (email permission, informal collegial
sharing, broadcast clips without an agreement, internal consented recording; 7 corpora). Legally
weakest tier, and flagged as such --- see Section~\ref{sec:limits}.
\end{enumerate}
One attribution-licensed corpus carries a narrower consent term than its licence:
UGAkan's participant consent form limits use of the recordings to training an ASR
model.

\subsection{Provenance and transfer chain}
\label{sec:provenance}

For every constituent corpus the register records: acquisition status, licence
type and SPDX or canonical URL, the specific agreement artefact (signed PDF,
email thread identifier, or public dataset card), the signature date where
applicable, redistribution and commercial-use permissions, the cross-border
transfer path, the data controller, and the custodian institution. Transfers
recorded include US$\rightarrow$UK (SAP), US$\rightarrow$UK$\rightarrow$NL
(HeyJay!), US$\rightarrow$NL (XRMB), NL$\rightarrow$UK (the Radboud and INT
Leiden corpora), CN$\rightarrow$UK (MDSC, CDSD), SK$\rightarrow$UK (EWA-DB), HU$\rightarrow$UK (the
Hungarian corpora) and CO$\rightarrow$MX$\rightarrow$UK (PC-GITA).

Consent provenance is documented at corpus level rather than speaker level and
its granularity varies with the instrument. For most corpora consent was obtained
by the originating institution under its own ethics approval, and the register
records the custodian and agreement rather than the underlying protocol. One
record carries more specific terms because it describes a single participant: one
Dutch pilot speaker recorded internally is covered by an SMF consent form. We do not hold, and therefore do not claim to publish, a
per-corpus ethics-committee mapping.

\paragraph{Re-identification risk for single-participant cohorts.} Four of
CANDOR's constituent records describe exactly one person each, and this warrants a
statement a corpus paper would not normally need to make.

The risk is not that dysarthric speech is unusually identifiable. \citet{tayebiarasteh2023}
report a mean equal error rate of 0.89\% for automatic speaker verification across
more than 3,800 pathological speakers, establishing speech as a working biometric
at scale, but find that dysarthria specifically yields results comparable to those
for healthy speakers, with dysphonia the exception. The risk for a cohort of one
arises from a different mechanism.

\citet{wiepert2024} show that re-identification risk is inversely related to the
size of the search space an adversary must consider: in their experiments the
ratio of false to true acceptances rose from 0.02 at $10^{5}$ comparisons to 1.41
at $6\times10^{6}$, with near-perfect precision at the small end. They note that
risk increases where metadata renders a population unique, and that a participant
known to have taken part in a shared-data study effectively reduces the search
space to one. We ran their published method on three of CANDOR's single-participant cohorts;
the fourth, a Hungarian recording, has not yet been assessed. We
reproduce the relationship: against a corpus-scale reference population of
$\approx$2{,}300 speakers, precision falls to 0.017--0.14 for two of the three
cohorts, while at a cohort-of-one search space it approaches unity. One cohort
departs from this pattern --- a single speaker who remained re-identifiable at
corpus scale (precision 0.82); we report this as an empirical outlier and do not
attribute it to aetiology, because with a single speaker it cannot be separated
from recording-channel or speaker-idiosyncratic factors, and the large-scale study
above finds dysarthria no more identifiable than healthy speech. The mechanism the
data support is therefore search-space size, not pathology: for a cohort of one,
ordinary corpus-description practice --- naming the country, the clinical
condition, the recruiting institution, the protocol and the agreement date --- is
jointly sufficient to identify the individual to anyone with access to that
institution's patient list, even though no single item is identifying alone.

The same work finds that the elicitation task modulates risk in a direction that
is not intuitive. Cross-task, connected speech is substantially easier to identify
than non-connected speech: sentence repetition yields a false-to-true ratio of
0.54 against 98.5 for vowel prolongation. Within task, this inverts, and vowel
prolongation becomes among the most identifiable, at 0.39. Their recommendation is
that non-connected recordings are safer to share only while no comparable public
dataset exists for matching. We note that this condition does not hold within
CANDOR: VOC-ALS comprises sustained vowels and diadochokinetic tasks and is
distributed through Synapse under the custodian's terms (non-commercial research
only, no redistribution), so a within-task reference set for that material
already exists outside CANDOR for researchers who obtain access under those terms. CANDOR records the elicitation task per corpus in the register, and the planned public register carries it
(Section~\ref{sec:comp}), which makes this assessable per corpus rather than in
aggregate, and we regard that as one of the practical arguments for recording it.

We have therefore reduced the identifying detail carried in this paper to the
minimum the scientific claims require: language and aetiology class. Institution,
country, protocol name, countersignature date and any within-corpus identifier are
held in the restricted part of the register under the same access control as the audio, and are released only to parties who
have already signed the relevant agreement.

Per-patient adapters for single-participant cohorts are excluded from the Tier 3
release of Section~\ref{sec:access}; the general-purpose model weights, trained
across the full corpus, are not, and the distinction follows the search-space
argument above rather than an assumption that adapters are inherently unsafe. The
re-identification assessment gating that exclusion follows the published method of
\citet{wiepert2024}, whose implementation is openly available, and we report its
result above rather than describing the assessment as forthcoming; no per-patient
adapter has yet been trained for any of these three speakers, so what we assess is
the re-identifiability of the audio itself, which is the governing risk.

One question remains open. Institutional ethics approval for a constituent corpus authorises the originating
study; it does not automatically authorise secondary aggregation into a further
corpus or the release of derived models. For the constituent corpora we hold the
custodian's agreement and the licence and rely on the custodian's own approval.

\subsection{Inheritance of source terms by derived artefacts}
\label{sec:inherit}

CANDOR produces three classes of downstream object from its constituent corpora,
and only a minority of the 26 source instruments address licence inheritance at
all --- the UIUC TDUA, which defines trained models as a category distinct from
the data, and the bilateral agreements executed on our own template, which grant
the equivalent. All of these bear on
trained parameters (Class B); no instrument addresses the content-bearing or
aggregate classes, and the great majority are silent on inheritance entirely. The
classes differ in how much participant-identifiable content
they carry, and CANDOR's policy differs accordingly. Collapsing them into a
single rule is both unusable and untrue of practice: it would forbid publishing
the aggregate measures on which the field depends.

\paragraph{Class A: content-bearing artefacts.} Source recordings, and
derivatives preserving the linguistic content and speaker characteristics of a
source recording. For these CANDOR applies a most-restrictive intersection
across five dimensions independently:

\begin{quote}
\emph{A Class A artefact inherits the intersection of the most restrictive terms
of every source that contributed to it, across every dimension independently:
redistribution, commercial use, derivative works, persistent storage, and
retention.}
\end{quote}

Dimension-independence is the operative part. A pool built from a CC-BY-NC
source and a no-cloud-storage DUA source is simultaneously non-commercial
\emph{and} storage-restricted, though neither source imposes both. The
alternative --- reasoning case by case about whether a voice-converted utterance
still constitutes the source recording --- produces inconsistent decisions and
cannot be audited. Derivative permission is listed as a dimension because our
own implementation originally omitted it, with the consequence documented in
Section~\ref{sec:derived}.

\paragraph{Class B: trained model parameters.} Fine-tuned ASR weights,
per-patient adapters and severity embedding models. These are learned
parameters, not recordings, and CANDOR does not treat their release as
redistribution of a corpus. The position is not merely ours. The data transfer
agreement governing the corpus that supplies the majority of CANDOR's dysarthric
speakers defines models as a distinct category: machine learning models trained
on the data, provided they include no more than a de minimis portion of it, are
\emph{Results} rather than \emph{Data}; the agreement states that the recipient
is encouraged to make Results publicly available, permits their use for
commercial purposes, and preserves that right after the agreement terminates.
The bilateral agreements executed on our own template grant the equivalent in
terms.

Two constraints follow, and both bind. First, the de minimis qualification is a
substantive test rather than a formality: a model that has memorised training
utterances and can emit them may cease to qualify. An extraction assessment is
therefore a precondition of release, not a courtesy. Second, not every
instrument permits it. One multi-site agreement restricts use to scientific
understanding and method evaluation, excluding clinical or commercial
deployment; one single-institution agreement extends its prohibition to ``any
derived data \ldots\ for any purpose'', wording broad enough to reach a trained
adapter and drafted without reference to commerciality, so that non-profit
status does not answer it. Corpora under such terms are excluded from released
models unless written permission is obtained.

\paragraph{Class C: aggregate measures and methods.} Phonological
\dprime{} values, distributional statistics, corpus-level tables, analysis code
and methodology. These carry no participant-identifiable content and CANDOR
treats them as publishable without restriction. This is the class that makes the
programme's published work possible, and treating it as inheriting source
restrictions would make a governance-aware corpus an obstacle to the science it
exists to support.

\paragraph{Processing is not publication.} The storage dimension in Class A
restricts persistent storage and public availability, not transient
computation. CANDOR stages data to rented GPU infrastructure for training and
deletes it on completion; it does not place source audio in publicly accessible
storage under any circumstances.

We offer this as a reusable governance policy rather than a legal conclusion. It
is what CANDOR adopts, not an interpretation of the underlying instruments, and
for Class A it is deliberately stricter than several sources would individually
require.

\subsection{Reconciling an audit register at scale}

A practical observation for others maintaining comparable registers. During
preparation of this paper we verified the register's aggregate rows by summing
constituent records and found three stale subtotals and an understated
healthy-control grand total, caused by corpora appended without recomputation of
the aggregates. All figures in this paper derive from recomputed sums.
Aggregate rows in a manually maintained register should be treated as derived
quantities and recomputed on every write, not as stored values.

\section{Harmonisation}
\label{sec:harm}

\subsection{Audio and manifest standardisation}

All audio is converted to 16\,kHz mono 16-bit PCM. Every corpus is expressed as
a \texttt{manifest\_v2.jsonl} record set with eleven required fields:
\texttt{audio\_filepath}, \texttt{duration}, \texttt{text}, \texttt{speaker\_id},
\texttt{severity}, \texttt{language}, \texttt{dataset}, \texttt{content\_type},
\texttt{aetiology}, \texttt{is\_dysarthric} and \texttt{split}. An optional
\texttt{original\_severity} field preserves the custodian's label whenever
CANDOR overrides it, so that no original annotation is destroyed by
normalisation.

Two recurring practical problems are handled at manifest construction. Speaker
identifiers collide across corpora --- and, in corpora organised by recording
site, within them --- so identifiers are namespaced by originating subfolder.
Splits are speaker-disjoint by construction and verified by an auditor that
reports any speaker appearing in more than one split; this check is mandatory
before any training run, having previously caught silent leakage affecting
dozens of speakers.

\subsection{Why severity does not harmonise}

The fifteen corpora carrying severity labels use five mutually
non-commensurable clinical instrument families plus model-derived pseudo-labels,
listed in Table~\ref{tab:sev}. They measure
different constructs. Intelligibility scores measure listener comprehension.
UPDRS speech items rate clinician-perceived speech impairment. ABDAS and GRBAS
measure, respectively, self-assessed bulbar function and perceptual voice
quality. Perceptual ratings capture direct listener
or clinician impressions of severity. Two corpora carry model-derived
pseudo-labels rather than clinical ratings.

\begin{table}[htbp]
\centering
\small
\caption{Severity instruments across the 15 labelled corpora. The five clinical
instrument families and the model pseudo-label category are mutually
non-commensurable; no validated mapping between
them exists.}
\label{tab:sev}
\setlength{\tabcolsep}{4pt}
\begin{tabular}{p{3.4cm}rp{6.6cm}}
\toprule
\textbf{Instrument} & \textbf{Corp.} & \textbf{Corpora} \\
\midrule
Intelligibility & 8 & TORGO, UA-Speech, SAP (partial), COPAS, CHASING, Treasure Hunters 1, SSNCE Tamil, MDSC \\
UPDRS (speech) & 1 & PC-GITA \\
Model pseudo-label & 2 & IPVS, CDSD \\
Perceptual rating & 2 & HeyJay! (3-rater mean), SMF Dutch pilot recording \\
ABDAS & 1 & VOC-ALS \\
GRBAS & 1 & NeuroVoz \\
\bottomrule
\end{tabular}
\end{table}

\begin{figure}[htbp]
\centering
\includegraphics[width=\linewidth]{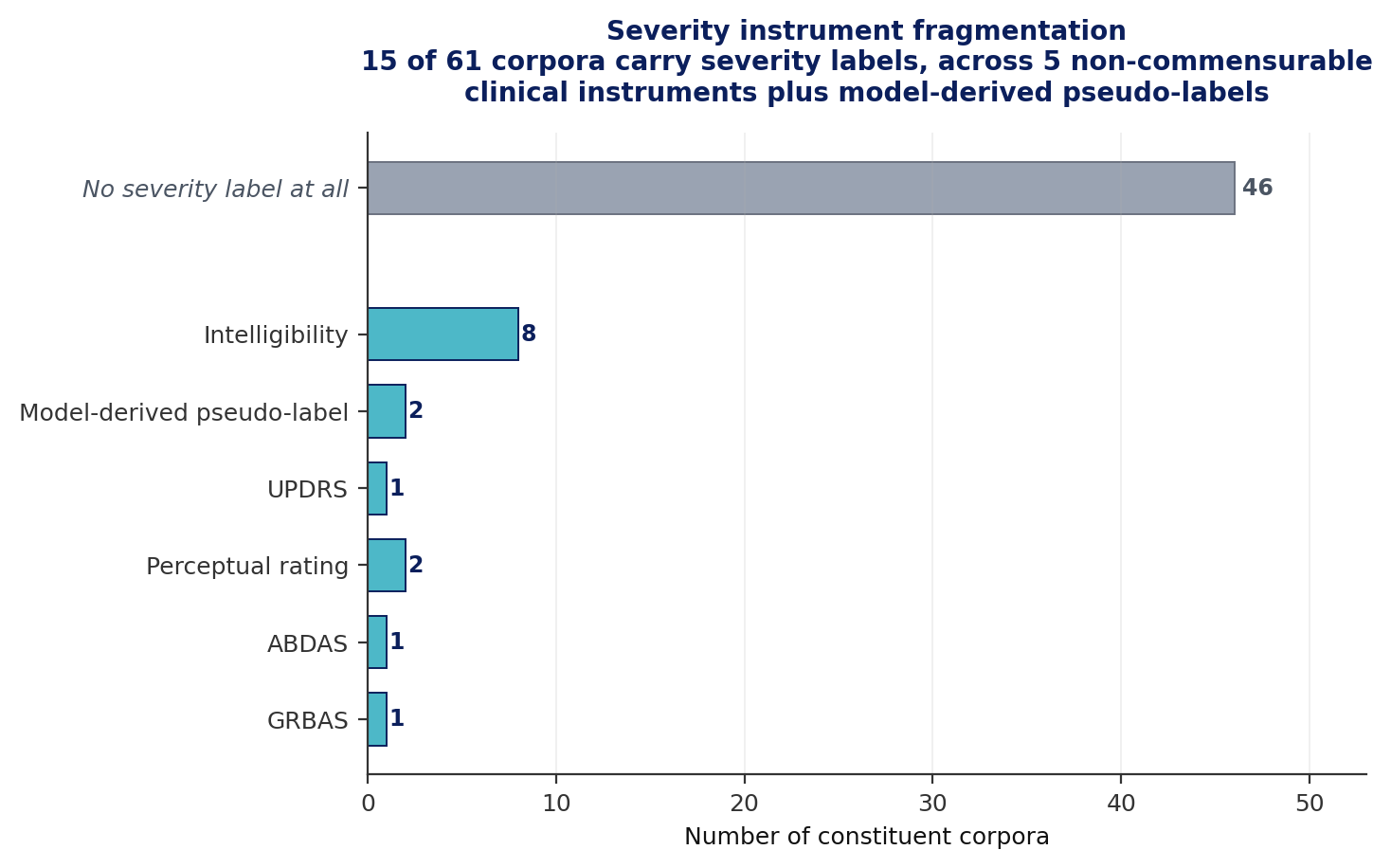}
\caption{Severity instrument fragmentation. Each bar is the number of
constituent corpora using a given labelling instrument. Fifteen of 61 corpora
carry severity labels, spread across five mutually non-commensurable clinical
instruments plus model-derived pseudo-labels; no instrument serves more than
eight corpora and three serve only one each. The grey bar is the 46 corpora
carrying no severity label at all, and it is longer than every labelled category
combined.}
\label{fig:severity}
\end{figure}

Figure~\ref{fig:severity} shows the resulting fragmentation. Even within an
instrument family, thresholds differ. The intelligibility group
alone contains a 1--7 clinician scale mapped to three bands (SAP), a 1--5
perceptual listener rating with a 3.5 mild boundary (CHASING), and
percentage-intelligibility bands (MDSC).
Mapping these onto a shared ordinal scale requires either speakers rated on
multiple instruments --- which we do not have --- or an assumption of
distributional equivalence across cohorts that differ in aetiology, language and
recording condition, which we are not willing to make.

\textbf{CANDOR does not provide a validated
cross-corpus severity mapping.} What it provides is the characterisation above,
a preserved original label for every rated speaker, and two
alternatives developed within the programme --- a training-free phonological
\dprime{} composite normalised to each corpus's own healthy-control
distribution (for Spanish, a pooled PC-GITA + NeuroVoz control reference), and contrastive severity embeddings trained per aetiology
(Section~\ref{sec:base}). Only the \dprime{} composite is label-independent: the
embeddings are trained with an ordinal triplet loss on clinical severity labels
and phonological pseudo-labels. The proposed path forward is ordinal anchoring against
a within-corpus healthy-control reference rather than absolute scale
translation, but we present that as a direction, not a validated result.

\subsection{Derived and synthetic pools}
\label{sec:derived}

CANDOR maintains 20 derived pools whose dysarthric portion contributes 639,113
utterances and 943.9 hours. Four further pools, the German and Spanish voice-converted and XTTS pools,
carry content derived from the OneVoice-MSD 2026 subsets and are excluded from
this paper with those subsets. These are \emph{never} counted as corpus scale: their speakers are
re-renderings of speakers already present in the real corpora, except that the
Japanese voice-converted pool derives from SS-JDSC, which is excluded from the
dysarthric inventory (Section~\ref{sec:comp}). The pools are used
for training-set augmentation. Section~\ref{sec:base} compares performance on
held-out synthesis and real speech; it does not isolate the effect of
augmentation.

The pools are produced by three mechanisms that differ in how much of the source
speaker survives, and the difference determines which inheritance class of
Section~\ref{sec:gov} applies. It is not a presentational detail: assigning a
single posture to all 20 would both overstate the restriction on some and
understate it on others.

The mechanism determines the class, but it does not by itself determine the
outcome. Recomputed under the split, 18 of the 20 pools remain
non-redistributable --- 5 of 7 voice-converted, all 8 model-generated, and
all 5 healthy-control pools. That the healthy-control pools are uniformly
restricted despite containing no dysarthric acoustics is the clearest
illustration of the point: it is their source transcripts that carry the
terms.

\paragraph{Voice-converted pools (7).} Each output is a real dysarthric
utterance whose voice timbre has been replaced by a healthy-control donor
through k-nearest-neighbour conversion. Transcript, timing, articulation and
prosody of the source speaker are preserved in full; only the voice is
exchanged. These are Class A without qualification. A voice-converted utterance
is the source recording in a different voice, and treating it otherwise would
allow voice conversion to launder restricted audio into an unrestricted pool.

\paragraph{Model-generated dysarthric pools (8).} Seven severity-matched XTTS-v2
LoRA pools and one F5-TTS pool. Here an adapter is fine-tuned on dysarthric
speech and then used to synthesise \emph{novel} utterances --- text drawn from
public sentence corpora or generated --- rendered in healthy-control donor
voices. No source utterance survives in the output. What the pool inherits is
not a recording but a learned characterisation of dysarthric articulation,
carried in the adapter, which is a Class B artefact. These pools remain subject to source instruments governing modelling. Their
release status also depends on the applicable terms for donor-voice recordings
and source text. The Class C definition in Section~\ref{sec:inherit} concerns
non-identifying aggregate measures; it does not, by itself, establish the status
of generated audio. For this paper, all generated audio is treated as
non-redistributable.

\paragraph{Healthy-control synthesis (5).} One Piper pool of 445,988 utterances
and four legacy sentence pools. These contain no dysarthric acoustics whatever:
source transcripts are synthesised in healthy voices, providing the matched
control side of paired training data. The only element inherited from a source
corpus is its text. The source transcripts carry restrictions that prevent
redistribution of these pools. A synthesiser's licence alone does not establish
the release status of outputs that also depend on restricted source
material.

\paragraph{Where our rule is stricter than the licence, and where we failed it.}
One derived pool fails the rule this paper proposes. The failure is against our
own policy, not against the source licence, and that distinction is the reason
the rule exists.

AVFAD is licensed CC-BY-NC-ND-4.0. The ND term is often read as prohibiting
derivative works outright; it does not. The licence grants the right to
``produce and reproduce, but not Share, Adapted Material for NonCommercial
purposes only'', where Share means providing material to the public. Creating a
derivative is permitted; distributing it is not. Our own rule is deliberately
stricter --- it treats an ND source as barring the derivative from being built at
all --- because a pool that exists but may never leave the institution is a
standing liability, and because the point at which a licence is read should be
before the compute is spent, not after.

The Portuguese pool was built anyway. It takes none of its dysarthric characteristics from AVFAD: those
come from severity-matched LoRA adapters trained on a separate corpus whose
agreement expressly permits model training. What it takes from AVFAD is
\emph{voices}. Synthesis is conditioned on speaker references drawn from 367
donors, of which 347 are AVFAD speakers --- 342 from its healthy subset and five
from its dysarthric subset --- with the remaining 20 from Common Voice. Cloning a
recording to condition the synthesis of new audio adapts that recording, so the
ND term is engaged through the voice-reference path even though the dysarthric
signal originates elsewhere.

The rule was applied consistently to
redistribution, commercial use, storage and retention, but derivative permission
was not among the four dimensions the pipeline checked, and the ND source entered
through the voice-donor slot rather than the dysarthric-source slot.

The pool has not been withdrawn. It remains in the corpus, and the Portuguese model
reported in Section~\ref{sec:base} was trained overwhelmingly on it: 18,239 of
that model's 18,273 training utterances are synthesis conditioned on these
references, against 34 utterances of real speech. The Portuguese rows in
Tables~\ref{tab:asr-synth} and~\ref{tab:asr-real} therefore rest on a pool our own
rule says should never have been built. The pool has never been shared, and under
the licence it may not be: it is held on institutional hardware, is excluded from
redistribution, and is barred from commercial use, so no term of AVFAD's licence
has been broken. We hold access to AVFAD from its custodian (University of Aveiro,
February 2026) and through TalkBank under CC-BY-NC-ND-4.0, which permits producing
adapted material for non-commercial use provided it is not shared; that is how
the pool is used. No separate written permission has been requested. We have
added derivative permission as a fifth dimension so that the omission cannot
recur. The Portuguese results should be read with this provenance in view. A
licence-inheritance rule is only as good as the enumeration of dimensions it is
checked against, and that enumeration should be published so that omissions are
visible.

\subsection{Use of generative AI}
\label{sec:ai}

The first author communicates exclusively via an AI-assisted eye-gaze interface
due to motor neurone disease. Generative AI (Claude, Anthropic) was used throughout
the research process: experimental design, code development, data analysis,
statistical computation, figure generation and manuscript drafting. All scientific
decisions, interpretations and conclusions were made by the authors, and no
AI-generated content was presented without author review. Corpus statistics
were computed by scripts from the dated register snapshot described in
Section~\ref{sec:comp}, with the assistance of Claude (Anthropic). The planned
deposit includes the scripts described in Section~\ref{sec:artefacts}.
A team member who did not take part in the computation independently re-derived
the critical figures from the dated register snapshot and checked the underlying
audio, including a full decode of the Ga recordings. The check reproduced the
counts, found duplicate recordings within constituent corpora, including five
PC-GITA files that repeat another task's take, and raised three
counting questions; the resulting corrections are those stated in
Sections~\ref{sec:comp} and~\ref{sec:limits}. The final totals, after those
corrections and the exclusion of the Japanese single-speaker corpus, were
recomputed by script from the dated register snapshot and the listed corrections;
that recomputation was not independently re-checked. The authors
retain responsibility for the scientific decisions, interpretations and final
manuscript.

\section{Baselines}
\label{sec:base}

The baselines below establish reference points on CANDOR rather than claiming
state-of-the-art results. A complementary line of work attacks the same
comparability problem from the opposite direction: PathBench
\citep{halpern2026pathbench} defines a unified intelligibility-assessment
benchmark over six public pathological-speech datasets and three evaluation
protocols, so that methods can be compared without each paper inventing its own
setup. PathBench standardises the \emph{protocol} on a fixed set of public
corpora; CANDOR addresses corpus assembly and governance upstream of that. The
two are complementary, and a natural future step is to run CANDOR's open subset
under PathBench's protocols.

\subsection{Scope}

Baselines cover seven languages: English, French, Hungarian, Italian, Dutch,
Portuguese and Mandarin. German and Spanish baselines are withheld because the
training data of their models included a corpus that is now excluded from
CANDOR (Section~\ref{sec:comp}); retraining them without it is left to subsequent work. The Qwen3-ASR-1.7B
base model \citep{qwen3asr} supports \textbf{30} languages, verified against the
\texttt{SUPPORTED\_LANGUAGES} constant in its released inference package; eleven
of CANDOR's seventeen fall inside that set. Seven are baselined here; German
and Spanish are withheld as above, and Czech and Japanese are not baselined
because neither has dysarthric speech in this release (Section~\ref{sec:comp}).
The other six --- Tamil, Slovak, Akan, Ga, Luganda and Swahili --- fall outside
the model's 30 and have no ASR baseline. The limit is the base model's language
coverage, not a property of our evaluation.

The scope decision is nonetheless itself a finding: \textbf{194} dysarthric
speakers in CANDOR --- Slovak 95, Akan 38, Swahili 24, Tamil 20, Ga 10 and
Luganda 7, led by a Slovak cohort that is the corpus's second-largest --- speak
languages that Qwen3-ASR, a current multilingual ASR model, does not cover
at all. That is the strongest available argument for why a corpus of this breadth
needs to exist.

\subsection{ASR}

The ASR baseline uses seven models: six per-language Qwen3-ASR-1.7B
fine-tunes trained with derived-pool augmentation, plus an English model
developed for the Speech Accessibility Project Challenge. Each per-language model
is trained in two epochs over different augmentation families --- \textbf{ep1} on
real speech plus $k$-nearest-neighbour voice conversion, \textbf{ep2} continuing
on real speech plus severity-matched XTTS synthesis --- and ep2 is the production
checkpoint. Where a contrast is needed below it is ep1 against ep2 within this
family; we do not report the superseded all-data multilingual predecessor, which
was trained on a different data composition and is not a controlled comparator
for anything. All six per-language models are, however, initialised from that
all-data checkpoint, and its training set of 1,027,832 utterances included 278
utterances (0.03\%) from the three OneVoice-MSD 2026 subsets that are excluded
from CANDOR (Section~\ref{sec:comp}). That training predates the exclusion. The
synthesis adapters used for augmentation were trained on SAP, TORGO and
LibriSpeech only.

\paragraph{What the evaluation manifest contains.} The held-out manifest for
each language carries three disjoint subsets, tagged per record:
\texttt{real\_only} (recordings of dysarthric speakers),
\texttt{knnvc\_disjoint} (speaker-disjoint voice conversions) and
\texttt{xtts\_heldout} (a held-out sample of the XTTS synthesis distribution,
voice-cloned from healthy-control Common Voice and MLS source speakers).

The two surfaces are reported separately and are not pooled:
Table~\ref{tab:asr-synth} gives the \texttt{xtts\_heldout} result,
Table~\ref{tab:asr-real} the \texttt{real\_only} result. Table~\ref{tab:asr-synth}
therefore measures performance on synthesised renditions of healthy voices, not
on dysarthric speech, and we label it as such.

Scoring the real subset required correcting compute-node paths carried in those
records, which did not resolve on the evaluation host; all audio paths were
verified to resolve before the figures reported here were computed. Before that
correction, the harness had dropped the real dysarthric audio in eight non-English
languages, and for a ninth, Japanese (whose single real speaker is
now excluded from the corpus), the audio was missing
(Section~\ref{sec:artefacts}; the defect, its fix and the pre- and post-correction
evaluation records are part of the planned deposit). The failure
mode is worth flagging for others running comparable evaluations, because it is
silent: a harness that discards what it cannot open, rather than failing, will
report a number for the wrong population with no outward sign that it has done
so.

The central finding of this section is that the two surfaces do not agree, and
that the gap is wide. Mandarin is the only language carrying both a synthesis
figure and a usable real-speech figure, and the same checkpoint scores
\textbf{4.72\% CER on held-out synthesis against 33.73\% on real dysarthric
speech} --- a factor of seven. Performance on the augmentation distribution is
not a proxy for performance on the population the corpus exists to serve, and
the synthesis table below should not be read as one.

\subsubsection*{Synthesis-distribution results}

Table~\ref{tab:asr-synth} gives the synthesis surface for the six non-English
languages. The English challenge-set result is real speech and is reported with
the real-speech results in Table~\ref{tab:asr-real}. The spread runs from 4.68\% to 38.55\% CER. It describes how well each model
transcribes the output of the pipeline that generated its own training
augmentation, and is reported for completeness and for comparability with the
real-speech table, not as a measure of dysarthric ASR.

\begin{table}[htbp]
\centering
\small
\caption{ASR performance on \textbf{held-out XTTS synthesis}
(\texttt{xtts\_heldout}), production ep2 checkpoint. These are \emph{not}
dysarthric-speech results: every source speaker is a healthy control from Common
Voice or MLS, re-rendered by CANDOR's own synthesis pipeline, and each language
draws exactly 20 such speakers. The column measures how well each model
transcribes the output of the augmentation pipeline it was trained on. All values
recomputed under the normalisation of Section~\ref{sec:base} (NFKC, casefold,
punctuation stripped). WER is suppressed for Mandarin, whose orthography
marks no word boundaries and for which whitespace tokenisation yields a tokenizer
artefact rather than a word error rate.}
\label{tab:asr-synth}
\setlength{\tabcolsep}{4pt}
\begin{tabular}{lrrr}
\toprule
\textbf{Lang} & \textbf{$n$} & \textbf{CER} & \textbf{WER} \\
\midrule
zh & 3,000 & 4.72  & --- \\
nl & 3,000 & 4.68  & 8.38 \\
hu & 3,000 & 16.61 & 33.64 \\
fr & 3,000 & 19.40 & 25.02 \\
it & 1,575 & 25.90 & 40.35 \\
pt & 2,205 & 38.55 & 53.25 \\
\bottomrule
\end{tabular}
\end{table}

\subsubsection*{Real dysarthric speech}

Table~\ref{tab:asr-real} reports the real-speech surface: every cell for which a
number can be computed, each with an explicit validity flag. Only the Mandarin row is both clean and sampled from more than a handful of speakers, and
three languages have no real-speech number, because none has held-out real speech. The English row is real speech from held-out
Speech Accessibility Project speakers, but it comes from a different model and
protocol, and the set is atypical rather than purely dysarthric speech (it
includes speakers with Down syndrome); it is not comparable to the other rows.

\begin{table}[htbp]
\centering
\small
\caption{ASR performance on real speech. Non-English rows use the production ep2
checkpoint and the normalisation used for Table~\ref{tab:asr-synth}. The English
challenge result uses a different model and protocol on held-out SAP speakers
with atypical speech and is not directly comparable. The confidence intervals
shown for non-English results are percentile intervals from 2,000 bootstrap
replicates resampling speakers; no interval is computed for a single-speaker
sample. The Validity column identifies overlap, sample-size and population
limitations.}
\label{tab:asr-real}
\setlength{\tabcolsep}{3pt}
\begin{tabular}{lrrrlp{4.6cm}}
\toprule
\textbf{Lang} & \textbf{utt} & \textbf{spk} & \textbf{CER} & \textbf{95\% CI} & \textbf{Validity} \\
\midrule
zh & 7,492 & 7 & 33.73 & [16.83, 59.61] & usable \\
\midrule
pt & 12 & 2 & 21.58 & [13.70, 29.45] & 12 utterances \\
fr & 130 & 1 & 17.72 & not computable & one speaker \\
\midrule
hu, it, nl & --- & --- & --- & --- & no held-out real speech exists \\
\midrule
en & --- & --- & 9.91 & --- & not comparable: a different model (the SAPC2 challenge English model) scored on the challenge Test1 set of held-out SAP speakers, atypical rather than purely dysarthric speech; WER 14.03 \\
\bottomrule
\end{tabular}
\end{table}

\paragraph{Why three languages have no real-speech number.} Hungarian,
Italian and Dutch have no real-speech row because all of their real dysarthric
audio was placed in ep2 training. This was a deliberate data-availability
decision, not an accident and not a leak: those three languages hold too little
real dysarthric speech to fund a speaker-disjoint held-out set of usable size,
and the choice at build time was between training on it and evaluating on it. We
chose training. The consequence is that no real-speech generalisation estimate
exists for them, and any figure computed on their \texttt{real\_only} records
measures training-set recall rather than recognition --- which is why we report
no such figure rather than reporting one with a caveat. Rebuilding real held-out
sets for these three requires more in-language data, and is the first item on the
baseline roadmap.

\paragraph{We do not report an augmentation effect.} The comparison that would
support one --- per-language augmented fine-tunes against a multilingual
all-data predecessor --- changes two variables at once
--- the training data composition \emph{and} the per-language specialisation ---
and so cannot attribute the difference to augmentation. The controlled experiment, training
the same recipe with and without augmentation, has not been run, and we make no
augmentation claim until it has.

What can be said is narrower.
Within the augmented family, the ep1$\rightarrow$ep2 step --- the step that
introduces the XTTS synthesis pool --- produces large gains on the synthesis
held-out set (Hungarian $-15.8$ points) that do
not reappear on real speech. In Mandarin, the one clean language with a usable real-speech estimate, the real-speech effect of that step shows \textbf{no
detectable change}: $+0.11$ points CER, 95\% CI $[-0.27, +1.22]$, $p = 0.884$, against
$-3.28$ points on the synthesis set. The data cannot, however, rule out a change of
up to 1.5 points: without the one speaker who contributes 5,443 of the 7,492
utterances, the 90\% interval is $[-0.24, +1.50]$. The two languages where the step appears to
reach significance on real speech cannot carry the claim. Portuguese
($+13.90$ points, $p < 0.001$) rests on twelve utterances from two speakers and
coincides with a long-form repetition hallucination that also renders its
voice-converted subset uninterpretable. Italian ($+8.26$ points) is measured on
audio that is almost entirely inside the model's own training manifest --- 341 of
344 utterances --- from two speakers, so it estimates training-set recall rather
than generalisation, which is why Table~\ref{tab:asr-real} carries no held-out
Italian real-speech figure. The Portuguese and Italian comparisons do not
establish an effect on held-out real dysarthric speech. Mandarin shows no
detectable change, but the data cannot rule out a change of up to 1.5 CER
points. Preferential improvement on held-out synthesis is consistent with
adaptation to the synthesis distribution, but the comparison does not isolate
that explanation.

We are not the first to observe this pattern. \citet{mokgosi2025} synthesise
dysarthric speech with an independent pipeline --- SpeechT5 with x-vector speaker
embeddings, fine-tuned with LoRA --- and report a similar pattern: in-domain
augmentation cuts TORGO word error from 0.60 to 0.17, but the gain does not carry
across datasets, where error rises to between 0.96 and 4.69. They conclude that
synthetic data improves in-domain recognition while cross-dataset generalisation
remains unsolved. This provides another example of limited transfer from
synthetic training to other datasets; it does not establish the cause of the gap
in our evaluation.

\paragraph{Repetition in Portuguese outputs.} Portuguese outputs exhibit
repetition on long-form input, and its voice-converted subset produces character
error rates above 100\%. We exclude those figures from the real-speech
generalisation comparison. The change between ep1 and ep2 does not establish
whether the cause lies in the training data, the training procedure or decoding
behaviour. Portuguese also has only seven dysarthric speakers and 0.4 hours in
the inventory.

\begin{figure}[htbp]
\centering
\includegraphics[width=\linewidth]{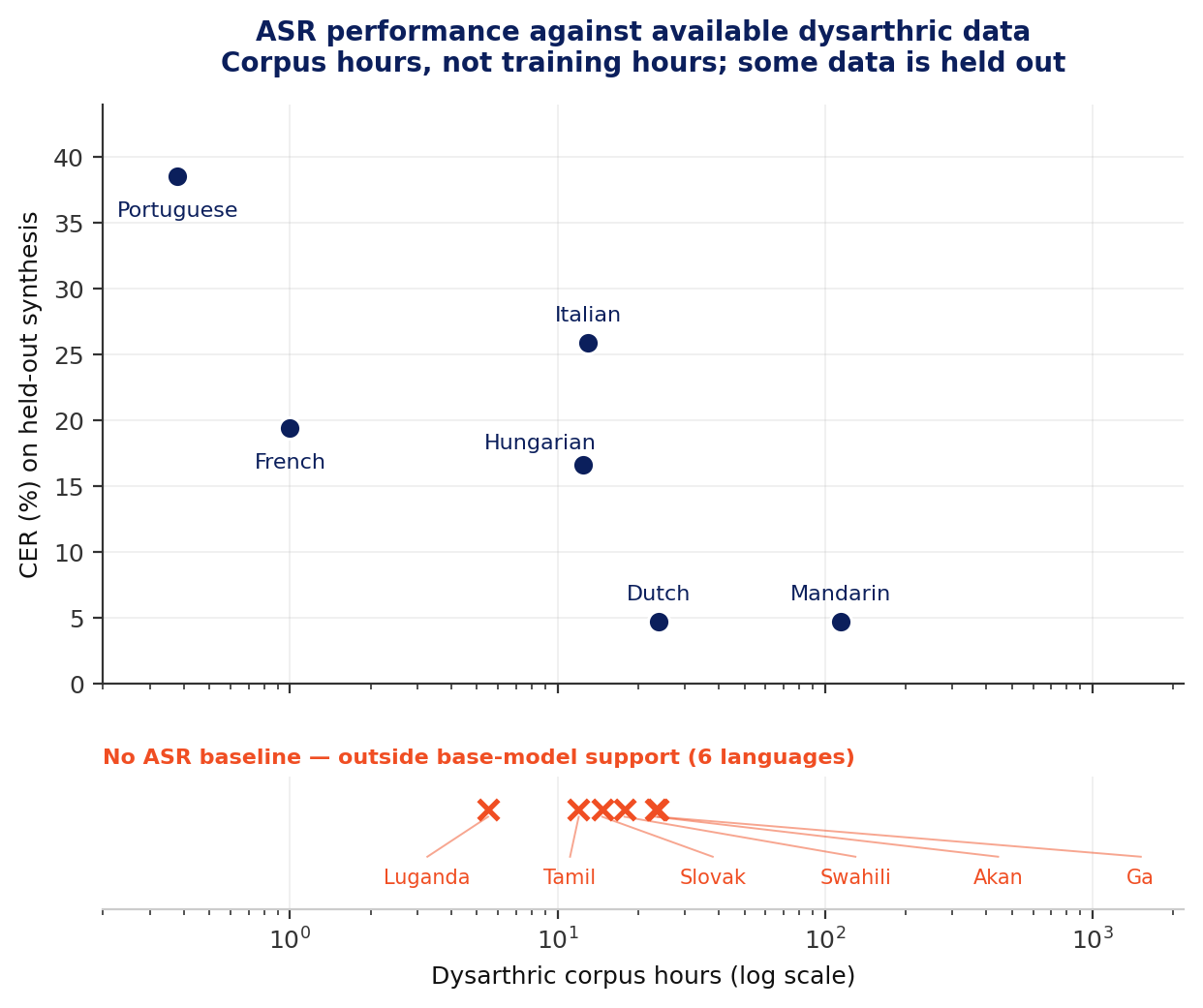}
\caption{Character error rate against available dysarthric data. Horizontal axis
is \emph{corpus} hours per language on a log scale --- the dysarthric speech
CANDOR holds, not the hours used in training, which differ because data is held
out. Upper panel: the six non-English baselined languages on the held-out
synthesis distribution (Table~\ref{tab:asr-synth}); German and Spanish, whose
baselines are withheld, are not shown. Lower strip: the six languages outside
the base model's supported set (Tamil, Slovak, Akan, Ga, Luganda, Swahili), which
have no CER at all and are given their own axis rather than being plotted at
zero. Note that several of them hold more dysarthric speech than languages in the
upper panel. \textbf{This figure plots synthesis performance, not dysarthric-speech
performance}; the real-speech evidence is in Table~\ref{tab:asr-real} and is too
sparse to plot.}
\label{fig:cer}
\end{figure}

\begin{figure}[htbp]
\centering
\includegraphics[width=\linewidth]{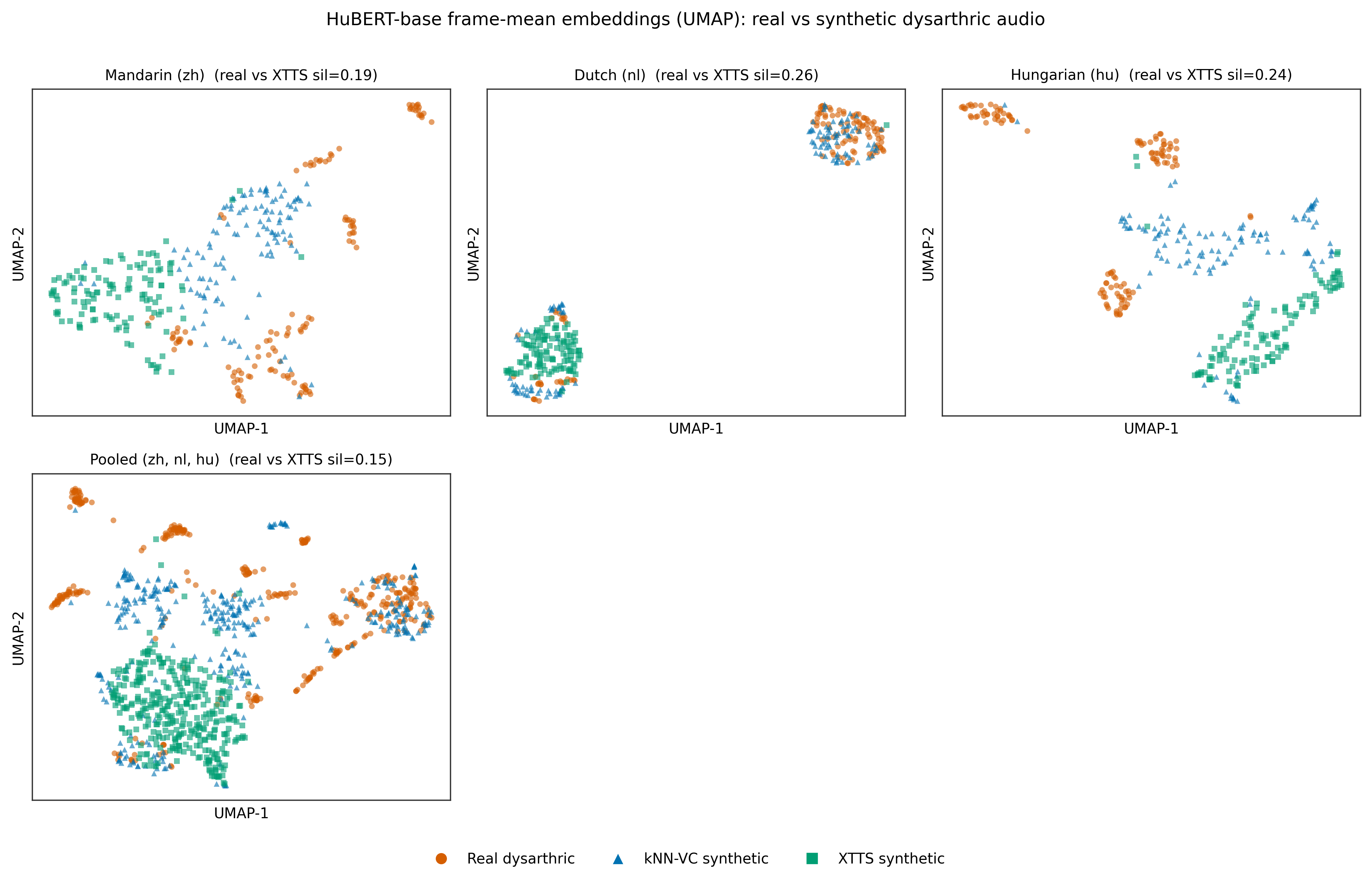}
\caption{HuBERT-base frame-mean embeddings of real and synthetic speech. UMAP
projections are shown by language and pooled, with real-versus-XTTS silhouette
scores in the panel titles, for Mandarin, Dutch and Hungarian. The pooled
silhouette is 0.03 for the voice-converted pool and 0.15 for XTTS; per-language
XTTS scores range from 0.19 to 0.26. Spanish is omitted because its synthesis
pools are excluded from this paper (Section~\ref{sec:derived}). No significance or equivalence test is reported. The comparison uses
training synthesis pools rendered in healthy-control donor voices
(Section~\ref{sec:derived}), rather than the held-out evaluation slice. A
$t$-SNE projection shows a similar pattern.}
\label{fig:embed}
\end{figure}

\paragraph{Embedding overlap between real and synthetic speech.}
Figure~\ref{fig:embed} compares HuBERT embeddings of real dysarthric speech with
the two synthesis families used for augmentation. The voice-converted pool
overlaps the real-speech embeddings closely (pooled silhouette 0.03). The XTTS
pool is more separated (pooled silhouette 0.15, reaching 0.26 in one language).
These analyses describe the relationship between the datasets in this
representation. They do not establish clinical equivalence or identify the
mechanism behind the difference between synthesis and real-speech performance.

\paragraph{The spread across languages is not a language-difficulty result.} A
seven-language evaluation on a consistent protocol might be expected to reveal a
range of language difficulty. It does not. The synthesis spread in
Table~\ref{tab:asr-synth} --- 4.68\% to 38.55\% CER --- is a spread in how well
each model transcribes its own synthesis pipeline, which is a property of the
augmentation regime and the base model's coverage, not of the languages. On real
speech we have one clean, usable estimate, and even within it, corpus
dominates language: Mandarin's two constituent corpora differ by
\textbf{2.3$\times$} (MDSC 15.83 against CDSD 36.31, both under the same model
and the same protocol; Table~\ref{tab:zh-corpus}), a ratio comparable to much of the variation between
languages elsewhere in the table. Recording conditions, elicitation task,
severity mix and speaker count all vary with corpus, and with a single clean
language there is no way to separate corpus from language. We therefore state
what the data support: \textbf{per-corpus difficulty on real dysarthric speech
varies widely, our real-speech evidence is concentrated in Mandarin at 33.73\%
CER (7,492 utterances, 7 speakers, 95\% CI [16.83, 59.61]), and the remaining
languages are either too small, single-speaker, or have no held-out real speech,
and are reported for completeness rather than as difficulty estimates.} That
confidence interval is itself wide: seven speakers pin real-speech performance
only to roughly the 17--60 CER band.

\begin{table}[htbp]
\centering
\small
\caption{Mandarin real-speech CER by constituent corpus, production ep2
checkpoint, same normalisation as Table~\ref{tab:asr-real}. CER is total
character edits over total reference characters, so the pooled figure weights
each corpus by its reference characters, not by its utterances.}
\label{tab:zh-corpus}
\begin{tabular}{lrrrrr}
\toprule
\textbf{Corpus} & \textbf{utt} & \textbf{spk} & \textbf{Ref.\ chars} & \textbf{Edits} & \textbf{CER} \\
\midrule
CDSD & 6,784 & 3 & 32,649 & 11,854 & 36.31 \\
MDSC & 708 & 4 & 4,693 & 743 & 15.83 \\
\midrule
Pooled & 7,492 & 7 & 37,342 & 12,597 & 33.73 \\
\bottomrule
\end{tabular}
\end{table}

Figure~\ref{fig:cer} plots the synthesis result against the dysarthric corpus
hours available per language, and makes the six unbaselined languages visible as
absence rather than omission. A per-corpus rather than per-language results table
is what this evaluation actually measures, and is the form the next revision of
these baselines should take.

\subsection{Severity embeddings}

Contrastive severity embeddings trained per aetiology \citep{muller2026paper3} (HuBERT-base backbone \citep{hsu2021hubert} with
an MLP projection to a 256-dimensional L2-normalised embedding, trained with a
three-stage contrastive objective) yield the speaker-disjoint held-out
linear-probe macro-F1 scores in Table~\ref{tab:sev-embed}. Per-aetiology
training outperforms a mixed-aetiology baseline in all three cases, by 22.6\% to
40.0\% relative. These are per-utterance scores from a single training run per
model. Each per-aetiology test set contains the target aetiology plus
language-matched healthy controls, speaker-disjoint from training: 49,755
utterances for CP, 36,325 for PD and 25,217 for ALS. Table~\ref{tab:sev-embed}
reports these original held-out per-utterance results; it carries no intervals.
They are summarised from \citet{muller2026paper3}, not recomputed for this paper;
the PD data of that study came in part from the OneVoice-MSD 2026 subsets, used
under that study's data agreement, which are excluded from the corpus described
here.
A separate analysis \citep[supplementary S.11]{muller2026paper3} re-scored the
retained test embeddings at speaker level under a fivefold cross-validation
within those test sets. It is not a restated held-out result. In it the test
sets hold 60 CP speakers (9 severe, 9 moderate), 104 ALS speakers (5 severe, 7
moderate) and 212 PD speakers (4 severe, 9 moderate), and its speaker-bootstrap
95\% intervals on speaker-level macro-F1 are approximately [0.30, 0.73] for CP,
[0.30, 0.62] for ALS and [0.32, 0.48] for PD. The ALS and PD test sets are 94.1\% and 66.7\% SAP; the CP test set
is 43.1\% CDSD. The ordering CP $>$ ALS $>$ PD is therefore directional, and the
results are largely English for ALS and PD.

\begin{table}[htbp]
\centering
\small
\caption{Per-aetiology contrastive severity embeddings, held-out speaker-disjoint
linear probe, unweighted, per-utterance macro-F1: the original held-out results,
single training run per model. Mixed column is a single model trained across all
aetiologies and evaluated on the same held-out subset. Results from
\citet{muller2026paper3}. The speaker-level intervals given in the text come from
a separate fivefold speaker-level re-scoring and do not apply to these values.}
\label{tab:sev-embed}
\begin{tabular}{lrrrr}
\toprule
\textbf{Aetiology} & \textbf{Per-aet.\ F1} & \textbf{Mixed F1} & \textbf{$\Delta$} & \textbf{Rel.} \\
\midrule
CP  & 0.829 & 0.676 & $+0.153$ & $+22.6\%$ \\
ALS & 0.788 & 0.596 & $+0.192$ & $+32.3\%$ \\
PD  & 0.715 & 0.511 & $+0.204$ & $+40.0\%$ \\
\bottomrule
\end{tabular}
\end{table}

\subsection{Phonological \dprime{}}

A training-free alternative projects frozen self-supervised embeddings onto
phonological contrast directions estimated from healthy-control speakers,
yielding per-speaker \dprime{} scores for contrast preservation
\citep[arXiv:2604.10123]{muller2026paper1}; \citep[arXiv:2604.21706]{muller2026paper2}. In the 890-speaker
validation of \citet[arXiv:2604.10123]{muller2026paper1}, all five consonant contrasts correlate
with severity ($\rho = -0.47$ to $-0.55$, $p < 10^{-48}$), and nasality
\dprime{} decreases monotonically from control to severe in six of seven corpora
tested. In \citet[arXiv:2604.21706]{muller2026paper2}, profile shapes are similar across
languages for PD, CP and ALS (mean cosine 0.978--0.987), but random directions
also give high cosines (about 0.93) and a prespecified test of
aetiology-specific cross-language stability was inconclusive; composite values
agree across six self-supervised encoders (inter-model $\rho = 0.83$--0.99).

Because this measure requires no severity labels and normalises to each corpus's
own healthy-control distribution (for Spanish, a pooled PC-GITA + NeuroVoz control
reference), it is currently CANDOR's most portable
severity proxy --- which is a statement about the scarcity of labels
(Section~\ref{sec:harm}) as much as about the method.

The speaker counts of the two works measure different populations: the 3,374
speakers analysed in \citet{muller2026paper2} include 1,426 healthy controls
across 25 datasets, whereas the 2,057 reported here are dysarthric speakers only.

\section{Access Model}
\label{sec:access}

CANDOR is not a deposit. The redistribution structure in
Section~\ref{sec:gov} makes a single-download release impossible: 19 of the 31
constituent corpora that carry dysarthric speech forbid redistribution outright,
and those corpora hold 86.2\% of the dysarthric speakers. Attempting a unified deposit would either exclude
the clinically valuable majority or breach the agreements under which it was
obtained.

We therefore adopt a three-tier model.

\paragraph{Tier 1 --- the register.} The planned public register contains
corpus name, language, aetiology, composition counts, elicitation task, severity
instrument, agreement type, licence, redistribution and commercial-use terms,
and canonical citation. It names custodians except for single-participant
records. For those records, custodian, institution, country and protocol
details remain restricted. Signature dates and transfer routes are also held in
the restricted register under the same access controls as the audio. The public
register is planned for release under CC-BY-4.0
(Section~\ref{sec:artefacts}).

\paragraph{Tier 2 --- access through custodians.} CANDOR does not sublicense
restricted corpora. Researchers must obtain access from the originating
custodian. We can route requests to that custodian, subject to the restrictions
on identifying information above; the access decision remains with the data
owner.

\paragraph{Tier 3 --- derived artefacts.} Where audio cannot be released,
derived measures often can. CANDOR may in future release model weights, contrastive
severity embeddings, phonological \dprime{} profiles, and per-patient adapters,
subject to the class scheme of Section~\ref{sec:gov} and to any source instrument
that governs modelling. These carry the analytical content
of the corpus without carrying the recordings, and for a substantial class of
downstream work they are sufficient.

\subsection{Artefact availability}
\label{sec:artefacts}

We state concretely what is available at submission, what is planned, and what
is conditional, and we use \enquote{released} only for what has been deposited.

\paragraph{At submission.} Nothing is yet deposited. At submission the
corpus-level description is this paper itself, including
Tables~\ref{tab:lang}, \ref{tab:aet}, \ref{tab:redist} and \ref{tab:sev}.

\paragraph{Planned.} Three artefacts are to be deposited in a public repository
(Zenodo) under CC-BY-4.0 after author sign-off, and cited by DOI in the final
version of this paper:
(i) the \textbf{public part of the constituent-corpus register} --- one row per
corpus with corpus name, language, aetiology, composition counts, elicitation
task, severity instrument, custodian (except for single-participant records),
agreement type, licence, redistribution and commercial-use terms, and canonical
citation, which is the
machine-readable form of Tables~\ref{tab:lang}, \ref{tab:aet} and
\ref{tab:redist};
(ii) the \textbf{harmonisation and audit code} --- the
\texttt{manifest\_v2.jsonl} schema definition and validator, the
speaker-disjointness auditor, the licence-inheritance calculator implementing the
five-dimension rule of Section~\ref{sec:gov}, and the scripts that compute every
count in this paper from the register; and (iii) the \textbf{evaluation split
specification} for every baseline reported here --- per-language, per-subset
lists of salted speaker hashes and utterance identifiers with durations and
reference-text hashes, sufficient for a third party to verify speaker
disjointness, subset membership and train/evaluation overlap without receiving
any audio or any transcript. The two-line defect that caused the evaluation
harness to skip unresolved audio paths silently, its fix, and the pre- and
post-correction evaluation records are included.

\paragraph{Conditional.} Model weights, contrastive severity embeddings,
phonological \dprime{} profiles and per-patient adapters would be released only where
the class scheme and the source instruments of Section~\ref{sec:gov} permit, and
trained parameters only after the extraction assessment that
Section~\ref{sec:gov} makes a precondition of release. No such assessment is
reported here, and no model is released with this paper.

\paragraph{What is not released, and why.} Audio is not released for any corpus, for the
reasons in Section~\ref{sec:gov}. Reference transcripts are not released where
the source licence forbids redistribution of derived text. Per-patient adapters
for single-participant cohorts are withheld pending the re-identification
assessment described in Section~\ref{sec:gov}. Training code for the ASR fleet
depends on a base model under its own licence and is released as configuration
and data-manifest specifications rather than as a redistributable bundle.

\paragraph{Persistent identification.} The constituent corpora retain their
original custodian identifiers where available. The planned Zenodo deposit will
identify the public artefacts listed above; it will not be an audio deposit.
Paper-specific contributions retain their own arXiv identifiers.

\section{Limitations}
\label{sec:limits}

\paragraph{Language skew.} English holds 1,325 of 2,057 dysarthric speakers, and
SAP alone holds 1,236. Excluding SAP, CANDOR is an 821-speaker corpus.
Cross-lingual claims should be checked against the per-language counts in
Table~\ref{tab:lang}, several of which are pilot-scale. The non-English
cohort is 732 speakers across fourteen languages.

\paragraph{Healthy-control imbalance and task mismatch.} 19,538 healthy-control
speakers against 2,057 dysarthric ones, and the control pool is overwhelmingly
read and parliamentary speech (LibriSpeech, VoxPopuli, Common Voice). It is
language-matched but not task-matched, not recording-condition-matched, and not
demographically matched to the clinical cohorts. Control comparisons drawn
across corpora rather than within them will be confounded.

\paragraph{Two aetiologies do not support cross-lingual claims.} Down Syndrome
comprises 166 speakers, 165 of them from SAP English; Stroke comprises 111, of
which 97 are from SAP and 102 English in total. Both are effectively single-corpus and
single-language. No cross-lingual or cross-corpus claim about either aetiology
is supportable from CANDOR, and we make none. Cerebral palsy, spanning seven
languages, is the only aetiology for which the corpus provides a broad
cross-lingual basis; Parkinson's disease is well-populated but almost
exclusively European (694 of 695 speakers).

\paragraph{Aetiology sparsity.} Thirty-three of 90 language $\times$ aetiology cells
are populated (Table~\ref{tab:aet}). Most aetiology-by-language contrasts a
reader might wish to draw are not available.

\paragraph{Diagnosis versus symptom is audited in only one corpus.} CANDOR
reports symptom-level speaker counts where symptom-level labels exist
(Section~\ref{sec:comp}), but only VOC-ALS publishes the speaker-level severity
scores needed to verify that the two conventions have been kept distinct. For
the remaining constituent corpora we rely on the custodian's own classification.
Where an aetiology presents variably --- the PD cohorts in particular --- a
diagnosis-level count reported as a dysarthric-speaker count would inflate that
corpus's contribution, and we cannot currently rule this out for corpora that do
not publish per-speaker severity. This is a limitation of the source
descriptions rather than of our processing, but it bounds the precision of the
2,057 figure.

\paragraph{Severity labels.} 15 of 61 corpora carry severity labels: 423
speakers in the 14 fully labelled corpora plus 188 in the SAP subset, 611 in
total, rated on five clinical instrument families plus model pseudo-labels. No
validated cross-corpus mapping exists (Section~\ref{sec:harm}).

\paragraph{Thin languages.} French is three speakers in the register, all from
public broadcast media (four in the manifests, which label one speaker dysarthric
whom the register lists as a control). Portuguese is seven speakers and 0.4 hours.
Japanese has no dysarthric speech in this release: its only clinical corpus,
SS-JDSC \citep{ogasawara2026ssjdsc}, records one speaker with velopharyngeal
insufficiency, which is not a dysarthria, and is excluded on the rule that
excludes COPAS's non-dysarthric speakers (Section~\ref{sec:comp}).

\paragraph{Unbaselined languages.} Six of seventeen languages, holding 194
dysarthric speakers, have no ASR baseline because they fall outside the 30
languages the underlying model supports (Section~\ref{sec:base}).

\paragraph{Real-speech baselines are essentially one language.} This is the
most severe limitation in the paper. Of seven baselined languages, exactly one
--- Mandarin --- supports a real-speech estimate that is both free of
train/evaluation overlap and drawn from more than a handful of speakers, and even
its 95\% confidence interval spans 17 to 60 CER. Three languages have no held-out
real speech at all, one
language rests on a single speaker, and one rests on twelve utterances. The German
and Spanish baselines are withheld (Section~\ref{sec:base}). No cross-lingual claim about dysarthric ASR difficulty is
supportable from these baselines and we make none (Section~\ref{sec:base}).

\paragraph{No controlled augmentation experiment.} The effect of derived-pool
augmentation on real dysarthric speech has not been measured under a controlled
design. The only usable real-speech test --- the ep1$\rightarrow$ep2 step in
Mandarin --- shows no detectable change ($p = 0.884$) but cannot rule out a
change of up to 1.5 points CER. Portuguese and Italian show apparent real-speech
changes across that step, but neither is interpretable: Portuguese rests on two
speakers and twelve utterances and shows long-form repetition
(Section~\ref{sec:base}), and Italian's real subset is almost entirely within
training. Claims of augmentation benefit made in earlier CANDOR material should
be read as claims about held-out synthesis only.

\paragraph{Demographics.} Age and sex are inconsistently available across
constituent corpora and are not centrally recorded in the register. Some
corpora supply full demographics, others none, and several required acoustic
inference to backfill sex labels. CANDOR cannot currently support
demographically stratified analysis at corpus scale.

\paragraph{Legally weak provenance.} Two corpora (French and German ALS clips
curated from publicly broadcast documentaries) have no agreement with the
broadcasters or with the speakers, and we claim no licence for them. They are
not redistributed. Their clips serve as held-out evaluation material and as
sources for derived augmentation pools (Section~\ref{sec:derived}). Separately, an unofficial
third-party copy of a German student project, with no speaker metadata or ethics
documentation, was used in German ASR training; it is not part of the corpus
described here and is excluded from every count in this paper.

\paragraph{Known-unreliable artefacts.} An externally sourced severity
classifier was applied to two corpora early in the programme and subsequently
found to perform below chance (13.3\% accuracy on TORGO). Its outputs are
retained only as a negative record and must not be treated as corpus labels.

\paragraph{Recordings shared across speaker identifiers within a corpus.} A
content fingerprint computed over every file of the corpora that contribute
dysarthric speakers or their own controls, with every candidate pair confirmed
by waveform cross-correlation, found recordings that appear under more than one
speaker identifier inside a single corpus. In UGAkan, 339 recordings are shared
across 9 identifiers. Three identifiers repeat another identifier's recordings
almost entirely (82 of 82, 96 of 98 and 15 of 16), so we count each as the same
speaker: UGAkan is reported as 38 speakers rather than 41 identifiers, and the
432 recordings (1.3 hours) that duplicate another recording in the corpus are
excluded from every count. One further UGAkan identifier labels unattributed
recordings, some of which also appear under two named speakers; we count it as
one speaker, which may misstate the number of people it holds. Elsewhere, 5
recordings are shared across 12 identifiers in EasyCall and one pair of
healthy-control identifiers in NeuroVoz; in VOC-ALS, a control and one of the 21
participants with an ALS diagnosis but no measurable impairment share all eight
of their recordings, and one of those recordings, a sustained vowel, also appears
under a third identifier counted as dysarthric, so the eight recordings link
three pairs of identifiers. Whether the identifiers in
these three corpora denote the same person can only be answered by the
custodians, so their speaker counts are the custodians' identifiers and may
overstate the number of distinct people. Separately, five PC-GITA
diadochokinetic files repeat, offset by up to 1.1 seconds, the recording held in
another task's file for the same speaker; each pair is counted once, so the five
repeating files are excluded from the utterance and hour counts (four dysarthric
and one healthy-control recording, 21 seconds in all).

\paragraph{Provenance records.} The excluded Japanese single-speaker corpus is
cited from its source publication \citep{ogasawara2026ssjdsc}; an earlier register entry that
left that publication unidentified has been corrected. The signed agreement for
PC-GITA is filed: the licence is signed 4 March 2026 by Prof.\ Antonio Ortiz Barra\~n\'on (Tec de Monterrey) with
the GITA laboratory at the Universidad de Antioquia.

\section{Roadmap}
\label{sec:roadmap}

Speech collection at one clinical site, the multidisciplinary ALS clinic at
TecSalud, Monterrey, started on 30 May 2026, with six ALS participants enrolled on
the first day; it contributes no data to the corpus described here. Other clinical
and industrial relationships are at letter-of-intent or in-discussion stage and
contribute no data.

Two expansions are in progress. The remaining 35 speakers of the ICPSR
atypical-speech corpus are pending an outstanding data request. African-language
coverage will expand through the existing CDLI relationship, which has already
contributed five dysarthric-bearing corpora across four languages.

\section{Conclusion}

CANDOR brings together 61 corpora across 17 languages and documents their
composition, provenance, access terms and limits. Its contribution is the
ability to examine a multilingual collection whose audio cannot be distributed
as a single resource.

In this collection, openly licensed corpora account for 10.5\% of dysarthric
speakers and 8.2\% of dysarthric hours. Excluding the Speech Accessibility
Project raises those shares to 26.2\% and 31.2\%. The relationship between
licensing and coverage therefore depends strongly on the composition of CANDOR;
it is not an estimate of the proportion of clinical speech available to the
field. Publishing corpus-level descriptions and permitted artefacts can make
such collections more inspectable without redistributing restricted audio.

Harmonisation exposed differences in speaker-counting conventions and severity
measurements, alongside errors in overlap detection and evaluation. The reported
baselines do not establish differences in ASR difficulty across languages.
Performance on held-out synthesis is not a substitute for evaluation on real
speech. Among the comparable non-English baselines, Mandarin supplies the only
real-speech estimate meeting the stated evaluation criteria, and its uncertainty
remains substantial.

The next steps are speaker-disjoint evaluation on real speech in more languages
and controlled tests of augmentation. CANDOR supplies a documented starting
point for that work, while preserving the distinctions between assembled data,
supported comparisons and artefacts that can be shared.

\section*{Acknowledgements}

This work was supported by The Scott-Morgan Foundation and SMF Labs. We thank
the custodians of all 61 constituent corpora and of SS-JDSC, which we examined
and excluded, and in particular the clinical
teams and participants whose consent makes this resource possible. The CDSD corpus is used under a signed academic licence from Prof.\ Su-Jing Wang (\textcopyright\ Su-Jing Wang, no redistribution). Portions of this work were conducted under restricted-use
agreements that prohibit redistribution of the underlying audio.

\section*{Statements and Declarations}

\paragraph{Funding.} This work was supported by The Scott-Morgan Foundation and
SMF Labs.

\paragraph{Competing interests.} Bernard Muller, LaVonne Roberts and Leo Kaiya
are affiliated with The Scott-Morgan Foundation or SMF Labs; Antonio Ortiz
Barra\~{n}\'{o}n is affiliated with Tecnol\'{o}gico de Monterrey and
collaborates with The Scott-Morgan Foundation on the CANDOR programme. The
Foundation and SMF Labs develop SMF VoxAI, a
patient-facing assistive communication system that uses models trained on
constituent corpora described in this paper. Several constituent corpora share
authors with Project Euphonia, which is used as a comparator in
Section~\ref{sec:related}; CANDOR makes no claim about unpublished Euphonia
holdings. The authors declare no other competing interests.

\paragraph{Ethics approval.} CANDOR is an assembly of pre-existing corpora. Most
constituent corpora were collected under their originating institutions' ethics
approval, and the approving body is identified in each corpus's canonical
citation. The authors do not hold an approval record for the two corpora of broadcast-media clips
(Section~\ref{sec:limits}), nor for the unofficial copy of a student project that
was used in German ASR training and is excluded from the corpus. One Dutch pilot
speaker was recorded internally under an individual consent form. Prospective
speech collection at TecSalud, Monterrey (Section~\ref{sec:roadmap}), which
started on 30 May 2026, contributes no data to the corpus described here; its
approval references are held by the collecting institution, not by the authors.

\paragraph{Informed consent.} The authors hold participant-level consent records
for only a small subset of the constituent corpora. For the remaining corpora,
the available evidence is documented at corpus level through the custody,
permission and transfer records described in Section~\ref{sec:provenance}. This
paper does not independently establish consent for every participant. The
limitations and exceptions in the available ethics and consent records are
stated in Sections~\ref{sec:provenance} and~\ref{sec:limits}.

\paragraph{Data and materials availability.} CANDOR cannot be redistributed as
a single audio deposit: 19 of the 31 constituent corpora carrying dysarthric
speech prohibit redistribution. No audio or model is released with this paper,
and no supporting artefact has yet been deposited. The planned public release
comprises the public register, harmonisation and audit code, and evaluation
split specifications described in Section~\ref{sec:artefacts}. These are to be
deposited under CC-BY-4.0 after author sign-off, with a DOI cited in the final
version. Access to restricted corpora must be sought from their custodians;
CANDOR does not sublicense them. Public custodian information excludes
single-participant records. Other derived artefacts remain subject to the source
terms and assessment requirements described in Section~\ref{sec:inherit}.

\paragraph{Author contributions.} Bernard Muller: Conceptualization, Methodology,
Software, Formal analysis, Investigation, Data curation, Visualization, Resources,
Writing -- original draft. LaVonne Roberts: Conceptualization, Funding acquisition,
Project administration, Supervision, Writing -- review \& editing. Leo Kaiya:
Methodology, Software, Writing -- review \& editing. Antonio Ortiz
Barra\~{n}\'{o}n: Investigation, Data curation, Writing -- review \& editing.

\bibliographystyle{apacite}
\bibliography{paper8}

\end{document}